\documentclass[10pt]{article} 
\usepackage[preprint]{tmlr}

\usepackage{amsmath,amsfonts,bm}

\def\eqref#1{equation~\ref{#1}}

\def\1{\bm{1}}

\DeclareMathAlphabet{\mathsfit}{\encodingdefault}{\sfdefault}{m}{sl}
\SetMathAlphabet{\mathsfit}{bold}{\encodingdefault}{\sfdefault}{bx}{n}

\usepackage{hyperref}
\usepackage{url}

\usepackage{dsfont}
\usepackage{graphicx}%
\usepackage{multirow}%
\usepackage{amsmath,amssymb,amsfonts}%
\usepackage{mathrsfs}%
\usepackage[title]{appendix}%
\usepackage{xcolor}%
\usepackage{textcomp}%
\usepackage{manyfoot}%
\usepackage{booktabs}%
\usepackage{subcaption}
\usepackage{caption}

\usepackage{enumitem}

\usepackage{float}

\usepackage{relsize}

\setcitestyle{square,numbers}  

\renewcommand{\eqref}[1]{(\ref{#1})}

\newtheorem{prop}{Proposition}
\newtheorem{definition}{Definition}[section]
\newtheorem{theorem}{Theorem}

\newtheorem{property}{Property}
\newtheorem{remark}{Remark}%
\newtheorem{proof}{Proof}

\title{Convolutional Rectangular Attention Module}

\author{
\name Hai-Vy Nguyen \\
\addr Ampere Software Technology -- Institut de mathématiques de Toulouse \\
-- Institut de Recherche en Informatique de Toulouse
\\
\name  Fabrice Gamboa \\
\addr Institut de mathématiques de Toulouse
\\
\name  Sixin Zhang\\
\addr Institut de Recherche en Informatique de Toulouse
\\
\name  Reda Chhaibi\\
\addr Laboratoire Jean Alexandre Dieudonné, Université Côte d'Azur
\\
\name  Serge Gratton\\
\addr Institut de Recherche en Informatique de Toulouse
\\
\name Thierry Giaccone\\
\addr Ampere Software Technology
}

\def\month{MM}  
\def\year{YYYY} 
\def\openreview{\url{https://openreview.net/forum?id=XXXX}} 

\begin{document}

\maketitle

\begin{abstract}
In this paper, we introduce a novel spatial attention module that can be easily integrated to any convolutional network. This module guides the model to pay attention to the most discriminative part of an image.  This enables the model to attain a better performance by an end-to-end training. In conventional approaches, a spatial attention map is typically generated in a position-wise manner. Thus, it is often resulting in irregular boundaries and so can hamper generalization to new samples. In our method, the attention region is constrained to be rectangular. This rectangle is parametrized by only 5 parameters, allowing for a better stability and generalization to new samples. In our experiments, our method systematically outperforms the position-wise counterpart. So that, we provide  a novel useful spatial attention mechanism for convolutional models. Besides, our module also provides the interpretability regarding the \textit{where to look} question, as it helps to know the part of the input on which the model focuses to produce the prediction.
\end{abstract}

\section{Introduction}\label{sec:intro}

In the field of computer vision, convolutional neural networks (CNNs) \cite{lecun2015deep} have proven to be powerful for many tasks, such as classification \cite{krizhevsky2012imagenet,he2016deep,tan2019efficientnet,li2021survey} or regression \cite{niu2016ordinal,lathuiliere2019comprehensive,zhang2019nonlinear}. To improve the performance of CNN models, one can insert attention modules that enable the model to produce more discriminative features \cite{hu2018squeeze,woo2018cbam,park2018bam,fu2019dual}. The high-level idea is to help the model to pay more attention to the most important elements of inputs, so that the resulted outputs get more discriminative. This is similar to human visual behavior, where one is more likely to look only at the important elements related to a considered task. This is why the attention mechanism in CNNs has always been of interest to researchers. Attention in CNNs can be divided in two main directions: spatial attention and channel attention. Channel attention encourages the model to select the most import channels (so the most import features) for a given task. However, the feature captured by each channel is not clear and interpretable. On the contrary, the spatial attention encourages the model to pay attention to the most discriminative region in the image. Therefore, this attention mechanism is interpretable as we know exactly which part of the image the model focuses on. In our work, we develop a new spatial attention module. In some previous works, one also proposed to combine spatial and channel attention together (for example \cite{woo2018cbam,park2018bam}). However, the interaction between spatial and channel attention impact is very complex.  
In our work, we focus solely on spatial attention and so better understand its impact on the model performance.

In classical approaches for spatial attention, the attention map is generated in a position-wise fashion (\cite{woo2018cbam,park2018bam}). This means that, for each input, the attention value is considered position by position to generate the whole attention map.

However, according to our experiments on images containing a single focal object, this method tends to provide irregular attention maps. That is, the boundaries of the attention regions are not stable, making it very difficult to generalize to new examples. Based on this remark, our intuition is to constrain the attention region to be more regular and stable. Hence, we propose to constrain the attention map to have a rectangular support. Then, this predicted rectangular support will determine the attention values for the whole input, where positions outside the support are assigned with zero attention. This is inspired by the fact that one is more likely to pay attention to a continuous region than to individual pixels. The important question is how to make this differentiable for an end-to-end training. In this work, we develop an attention module that answers this question. Our method systematically outperforms the position-wise approach, confirming correctness of our intuition. The qualitative results also lead us to the same conclusion.

\noindent
Our contribution can be summarized as follows.
\begin{itemize}
    \item We propose a novel spatial attention module that can be easily integrated to any CNN for an end-to-end training. This is achieved by using smooth rectangle attention supports respecting the equivariance with respect to translation, rotation and scaling on input images.
    \item We derive theoretical insights to better understand the attention module in idealized settings. This allows us to evaluate how well the predicted attention maps can {\it catch up} with the ground-truth maps on new samples. Besides,  our insights also provide a better understanding of how the attention module can help to produce more separable features in the context of classification.
    \item We conduct quantitative and qualitative experiments which show that our method is more stable than the standard position-wise approach. We also find that our predicted attention maps can follow smoothly a sequence of transformations (composed of translation, rotation and scaling) thanks to equivariance constraints (Section \ref{sec:equi}).
\end{itemize}

\noindent
\textbf{Paper organization.} Section \ref{sec:related_work} gives an overview of related works. Section \ref{sec:preliminary} recalls the general framework for inserting a spatial attention module into CNNs according to standard approaches. Then, Section \ref{sec:our_method} introduces our method in detail. In Section \ref{sec:theoretical}, we derive theoretical insights to better understand the attention module. Finally, quantitative and qualitative experiments are conducted in Section \ref{sec:exp}.

\section{Related works}\label{sec:related_work}

Attention mechanism has always been an important subject in Deep Learning. Different attention techniques are widely used in Natural Language Processing (NLP) models. We can name some applications such as neural machine translation \cite{bahdanau2014neural}, image captioning \cite{lu2017knowing,xu2016showattendtellneural}, speech recognition \cite{chorowski2014end} or text classification \cite{liu2019bidirectional,lin2017structured}. More recently, the transformer model \cite{vaswani2017attention} also uses attention mechanism and achieves impressive results in many NLP tasks.

In the context of CNNs, we can summarize the two main attention methods, including  \textit{spatial attention} \cite{wang2018non, yuan2021ocnet, huang2019ccnet,zhao2018psanet} and \textit{channel attention} \cite{hu2018squeeze,yang2020gated,lee2019srm,cao2020global,wang2020eca}. One can also combine the two attention methods \cite{woo2018cbam,park2018bam,fu2019dual,yang2021simam}. The channel attention helps to focus on the most important channels, by assigning high attention values on these channels. In other words, the channel attention helps the model to focus on the most important features. However, the feature learned by each channel is not
clear. In contrast, spatial attention learns {\it where} to focus on. This is more interpretable, as we know on which part of the inputs the model pays more attention to have the prediction.

To have an appropriate attention values for each channel and each position, one needs to carefully take into account the global context of the inputs. This is addressed in many previous works by proposing long-range dependency modeling \cite{cao2020global,woo2018cbam,hu2018squeeze,zhao2018psanet,wang2018non}. All previously mentioned methods have different module designs to model the attention in different ways. 

However, what all these methods have in common is that, the attention map is generated by considering position by position, without a global control on the smoothness of the attention map.
This makes it less regular and so more difficult to generalize to new samples 
(see experiment results in Fig. \ref{fig:qualitative_oxford_pets} and \ref{fig:compare_layer}). This contrasts with our method where the attention support is constrained to be rectangular and therefore more stable (see theoretical justification in Section \ref{sec:error}). Indeed, instead of considering position by position, our method predicts a rectangular support that determines the whole attention map. At the same time, the spatial long-range dependency is guaranteed by the subsampling process used in our attention module (see Section \ref{sec:technique}).

\section{Preliminaries and framework} \label{sec:preliminary}
\begin{figure*}[ht]
\centering
\begin{subfigure}{0.4\textwidth}
    \includegraphics[width=\textwidth]{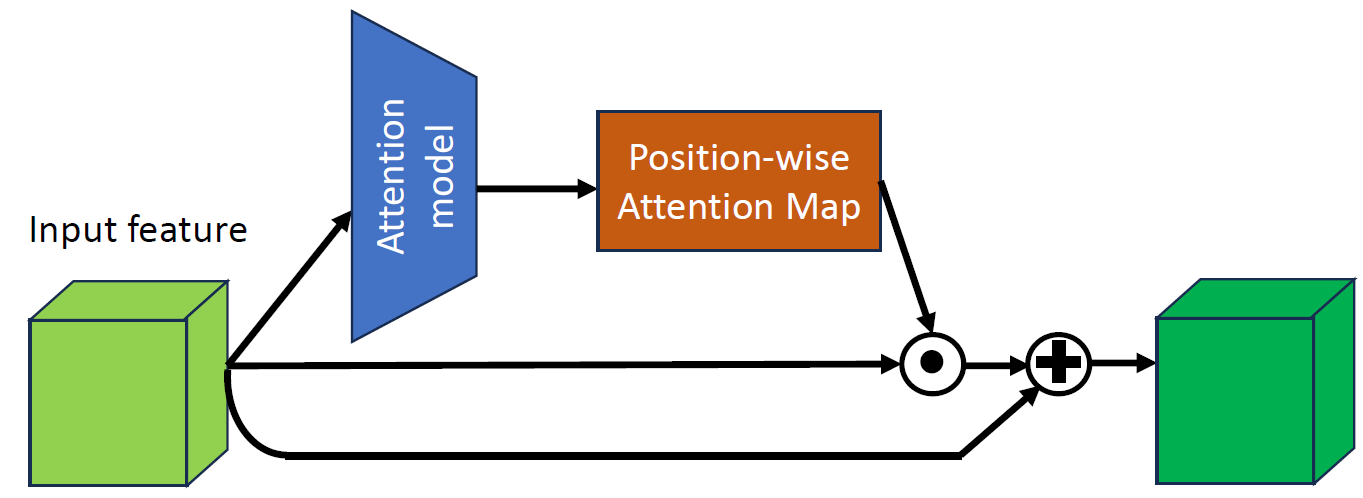}
    \caption{Spatial attention module with attention map generated in position-wise fashion in standard approaches.} \label{fig:classic_attention_module}
\end{subfigure} \hspace{0.05\textwidth}%
\begin{subfigure}{0.4\textwidth}
    \includegraphics[width=\textwidth]{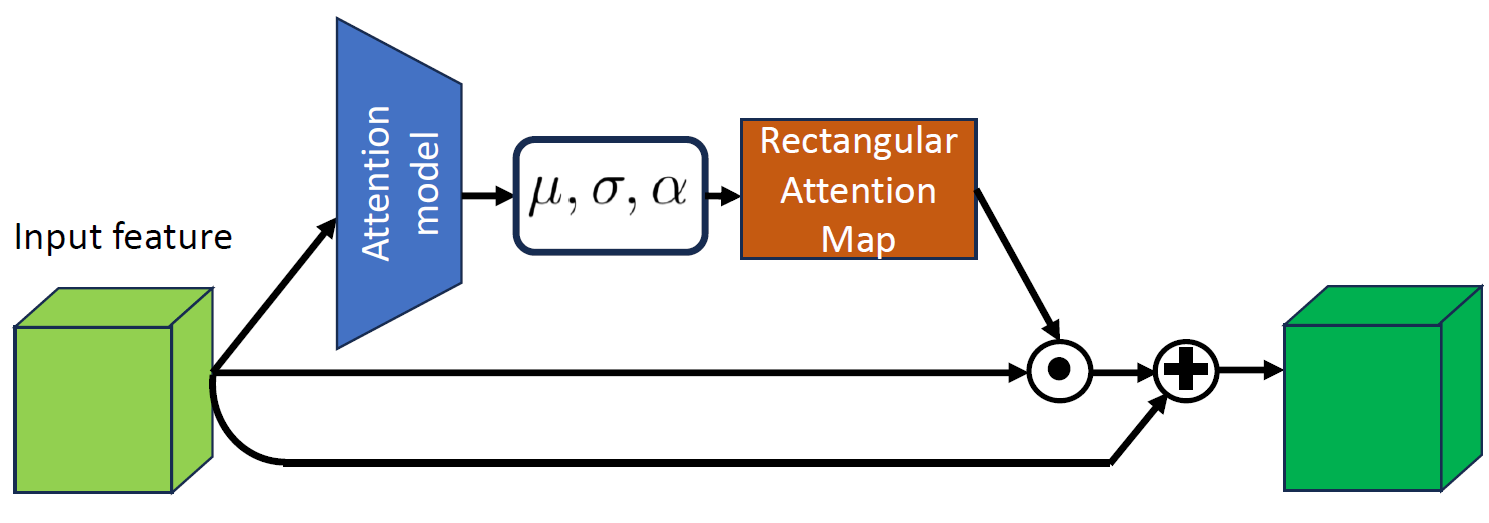}
    \caption{General scheme of our convolutional rectangular attention module.} \label{fig:rec_attention_module}
\end{subfigure}
\caption{General scheme of convolutional spatial attention modules in standard approaches (\ref{fig:classic_attention_module}) and our method (\ref{fig:rec_attention_module}).} \label{fig:attention_module}
\end{figure*}

In this section we recall some classical definition related to the spatial attention concept for convolutional neural networks.
To begin with, consider a task where the inputs are images. These images are fed into a convolutional neural network. Let us consider an intermediate layer that produces outputs of dimension $H\times W \times C$. $H$ and $W$ are the spatial dimensions (and so depend on the input dimension), and $C$ is the number of channels (independent of the input spatial dimension). Let $x\in\mathbb{R}^{H\times W \times C}$ denote such an output. We aim to learn a function $f$ valued in $[0,1]^{H\times W}$  that computes a spatial attention map associated with $x$. 
The purpose of the attention map is to highlight the important elements of $x$. For each position $(i,j)\in [H]\times[W]$, a new vector is calculated by multiplying $x(i,j)\in \mathbb{R}^C$ by its corresponding attention value $f(x)_{i,j}$. That is, all the channels in the same position is multiplied by the same spatial attention value $f(x)_{i,j}$. Let $f(x)\odot x$ denote this operation. 
It is usual to use the so-called residual connection, with the output $x+f(x)\odot x$ (for example in CBAM method \cite{woo2018cbam}), instead of $f(x)\odot x$.
%
This output is then propagated into the next layer of the neural network and follows the forward pass.
%
By training the model to optimize the performance of the desired task, the attention module learns to assign a high attention value to the discriminating parts of $x$ (attention values close to 1) and to ignore its unimportant parts (attention values close to 0). General scheme of this approach is depicted in Fig. \ref{fig:classic_attention_module}.

Usually, the attention map is generated in a position-wise fashion. That is, the attention for each position is computed by a module typically composed of spatial convolutional layers, 
or of pooling layers with weighted softmax.

Moreover, one can also combine spatial attention with channel attention for improving the model performance. In this work, we focus only on the spatial attention as it is more interpretable. In our experiment, we found that using a position-wise method often leads to attention map with irregular boundaries between important and unimportant parts of $x$.
%
Hence, in many cases, it harms the generalization capability of the model. That is, adding the position-wise attention module deteriorates the performance (compared to not using this latter module).

\section{Our method}\label{sec:our_method}
\subsection{A differentiable function with nearly-rectangular support} 
As discussed before, generating the spatial attention map in position-wise fashion could lead to irregular boundaries. Indeed, as the attention value for each position is computed separately, there is no guarantee for a regular and smooth attention map. This is confirmed in our experiment (Fig. \ref{fig:qualitative_oxford_pets} and \ref{fig:compare_layer}). Hence, we aim to constrain a map having a more regular shape. A natural choice for the support of attention maps is the rectangular shape. By moving the rectangle to cover the most discriminative part of the feature map, the model learns to focus on the most important part and to ignore non-relevant details. Furthermore, only  5 parameters are needed to characterize a rectangle, namely, its height, its width, its center position (2 parameters) and an orientation angle. Intuitively, by learning so few parameters (compared to standard position-wise approach), the attention module learns better. In addition, as the support of the attention map is rectangular, the model can generalize better for a new sample as it provides a more stable support. Our theoretical insights in Section \ref{sec:error} are aligned with this intuition.
Now, the question  is: {\it How can we build a differentiable module so that it can be integrated into any CNN model for an end-to-end training?}

\begin{figure}[ht]
\centering
\includegraphics[width=0.4\textwidth]{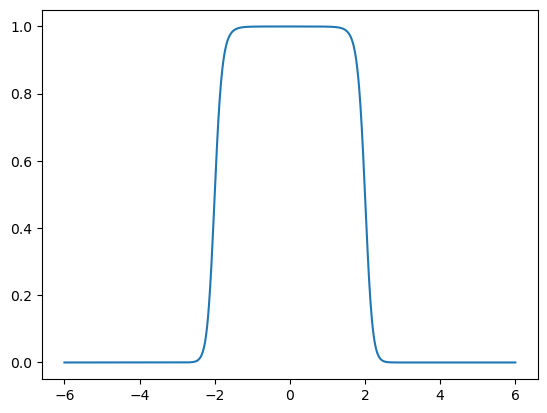}
\caption{{\it Window} function in Eq. \eqref{eq:scaled_sigmoid_function} with $s=10$, $\sigma=2$.}\label{fig:sigmoid_bell_function_main}
\end{figure}

The main idea is to construct a differentiable  function with values nearly $1$ inside a rectangle and nearly $0$ outside. 
To build this function, let us first consider the one-dimension case. In this case, we aim to build a function with values nearly $1$ in $[t_0-\sigma,t_0+\sigma]$ and nearly $0$ outside this interval. For this objective, we can use the following function
\begin{equation}\label{eq:scaled_sigmoid_function}
    \hat{\Lambda}_{s,t_0,\sigma}(t) =  \Lambda(s(1-(\frac{t-t_0}{\sigma})^2)) \ ,
\end{equation}
where $\Lambda(\cdot)$ is the sigmoid function, $s$ is a scale factor. An example of this function is shown in Fig. \ref{fig:sigmoid_bell_function_main}. To better understand why we have this function, please refer to Appendix \ref{sec:construct_bell_function}.

Now, we turn to the $2D$ case. We aim  to build a function with a rectangular support centered at $\mu = (\mu_1,\mu_2)$, having height $\sigma_1$ and width $\sigma_2$. Let us denote $\sigma = (\sigma_1,\sigma_2)$, tensorizing Eq. \eqref{eq:scaled_sigmoid_function} into $2D$ gives rise to,
\begin{equation}\label{eq:rec_function}
\hat{\Lambda}^0_{s,\mu,\sigma}(t_1,t_2) =  \Lambda(s(1-(\frac{t_1-\mu_1}{\sigma_1})^2)) \times \Lambda(s(1-(\frac{t_2-\mu_2}{\sigma_2})^2)) \ .   
\end{equation}

Now, to have more flexibility, we will add a rotation for this rectangle around its center $\mu=(\mu_1,\mu_2)$. This rotation operation in the plane writes,
\begin{equation}
\mathcal{R}_{\alpha}(t_1,t_2) =
\begin{pmatrix}
\text{cos}(\alpha) & \text{-sin}(\alpha) \\
\text{sin}(\alpha) & \text{cos}(\alpha) 
\end{pmatrix}
\begin{pmatrix}
t_1 - \mu_1 \\
t_2 - \mu_2
\end{pmatrix}
+\begin{pmatrix}
\mu_1 \\
\mu_2
\end{pmatrix}
\end{equation}

Note that if we want to rotate the support by an angle $\alpha$, we need to apply an inverse rotation, i.e., rotation by an angle $-\alpha$ for input of the function in Eq. \eqref{eq:rec_function}. Finally we work with the following collection of functions depending on 5 parameters $\begin{pmatrix}
    \mu_1\\\mu_2
\end{pmatrix}, \begin{pmatrix}
    \sigma_1\\\sigma_2
\end{pmatrix},\alpha$,

\begin{equation} \label{eq:final_function}
\hat{\Lambda}_{s,\mu,\sigma,\alpha}(t_1,t_2) =  \hat{\Lambda}^0_{s,\mu,\sigma}(\mathcal{R}_{-\alpha}(t_1,t_2)) \ .     
\end{equation}

\begin{figure}[ht]
\centering
\includegraphics[width=0.4\textwidth]{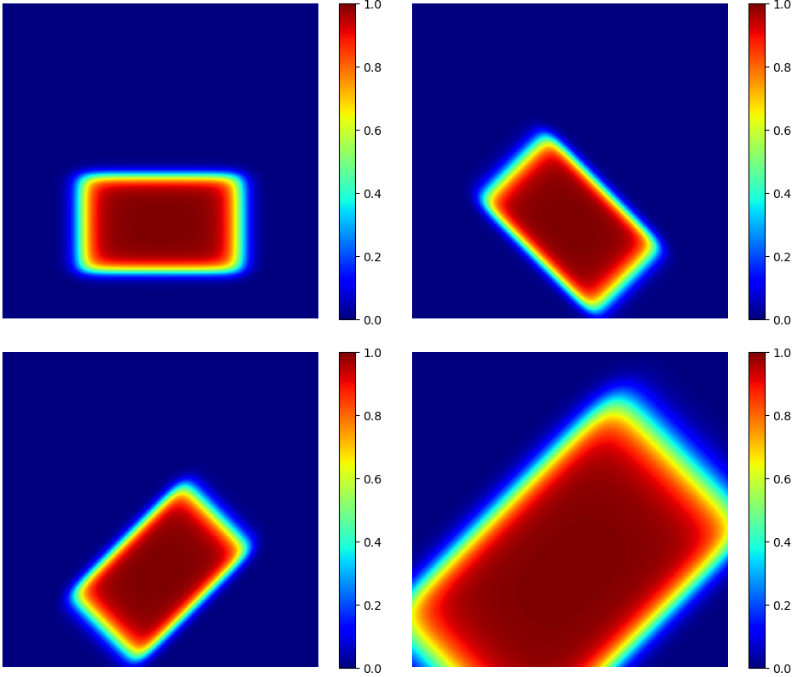}   
\caption{Illustration of a function with nearly-rectangular support in Eq. \eqref{eq:final_function}. The scaling factor $s=10$.} \label{fig:ex_rec_attention_intro}
\end{figure}

Some examples with different $(\mu,\sigma,\alpha)$ are depicted in Fig. \ref{fig:ex_rec_attention_intro}. We see that by changing the parameters of the function, we can easily move the support to capture different positions in the image.

\subsection{Convolutional rectangular attention module}
The general functional scheme of our method is shown in Fig. \ref{fig:rec_attention_module}. Let us consider an output $x\in \mathbb{R}^{H\times W\times C}$ of an intermediate convolutional layer. This output is the input feature of our module. This $x$ is fed into a trainable model to predict the parameters $(\mu,\sigma,\alpha)$. The attention map, denoted by $f(x)$, is generated by applying Eq. \eqref{eq:final_function} for each position $(i,j)\in[H]\times[W]$. Note that the coordinates of each position are rescaled in the range $[0,1]$.
That is, for each $(i,j)\in[H]\times[W]$,
\begin{eqnarray}
f(x)_{ij} = \hat{\Lambda}_{s,\mu,\sigma,\alpha}(\frac{i-1}{H-1},\frac{j-1}{W-1}) \ .
\end{eqnarray}
We also use the residual connection to obtain the final output feature as,
\begin{equation} \label{eq:attention_function}
\widehat{x} = x+f(x)\odot x \ .
\end{equation}

\noindent
\textbf{Rescaling - a standard step.} The idea of an attention module is to keep only important elements in the input and to  {\it drop} others (with an attention value $0$). In our case, the elements outside of the the predicted rectangular are dropped. Interestingly, this has a similarity with the regularization technique of {\it drop-out} \cite{srivastava2014dropout}. This last method consists in randomly dropping out certain activations during the training, by setting them to 0. In \cite{srivastava2014dropout}, it is argued that this can lead to a change in the data distribution. To overcome this drawback, the authors propose to rescale the output by the drop-out factor. Inspired by this remark, we also propose to rescale the obtained output of our module as follows,
\begin{equation}\label{eq:rescale_attention}
    \widetilde{f}(x) = f(x)\times \frac{HW}{\sum_{(i,j)\in [H]\times[W]}f(x)_{i,j}} \ .
\end{equation}
Here, $H\times W$ is the spatial size of $x$ (and so also the spatial size of $f(x)$). The intuition of this scaling is rather simple. Indeed, if $f(x)_{i,j}=1$ for all $(i,j)$ (no position is dropped), then we have $\widetilde{f}(x) = f(x)$. In contrast, if a portion of the initial input is dropped, then $\sum_{(i,j)\in [H]\times[H]}f(x)_{i,j}< HW$ (recall that $0\leq f(x)_{i,j}\leq 1$). Hence, $\frac{HW}{\sum_{(i,j)\in [H]\times[W]}f(x)_{i,j}}>1$, so that the scaled attention can compensate for the dropped ``portion". In some sense, this scaling helps to maintain a stable distribution of the output even when the rectangular support is varied in size. Hence, when this output is fed to the next convolutional layer to continue the forward pass, we have a better stability.
In our module, we apply this rescaling step.

\subsection{Technical points.}\label{sec:technique}
The attention module is typically composed of a few convolutional layers. In position-wise method, one needs to directly output an attention map with the same spatial dimension as input. Hence, one cannot reduce the spatial size after each layer in the module. Thus, to keep a reasonable running time, one cannot increase too much the channel dimension after each layer in this module. In contrast, in our method, we only need to output a vector of dimension 5, hence we can always reduce the spatial dimension by sub-sampling techniques (such as \textit{max pooling}) after each layer. Hence, we can increase significantly the channel dimension after each layer in the attention module, without increasing too much the running time. By increasing the channel dimension, the attention module learns richer features and produces more precise outputs. At the same time, applying sub-sampling for reducing spatial dimension after each layer in the attention module also allows us to better capture long-range spatial dependency. Finally, the implementation is very efficient, especially on GPUs, creating nearly no supplementary time at inference times. 

\subsection{Equivariance constraints} \label{sec:equi}

\begin{figure}[ht]
\centering
\includegraphics[width=0.8\textwidth]{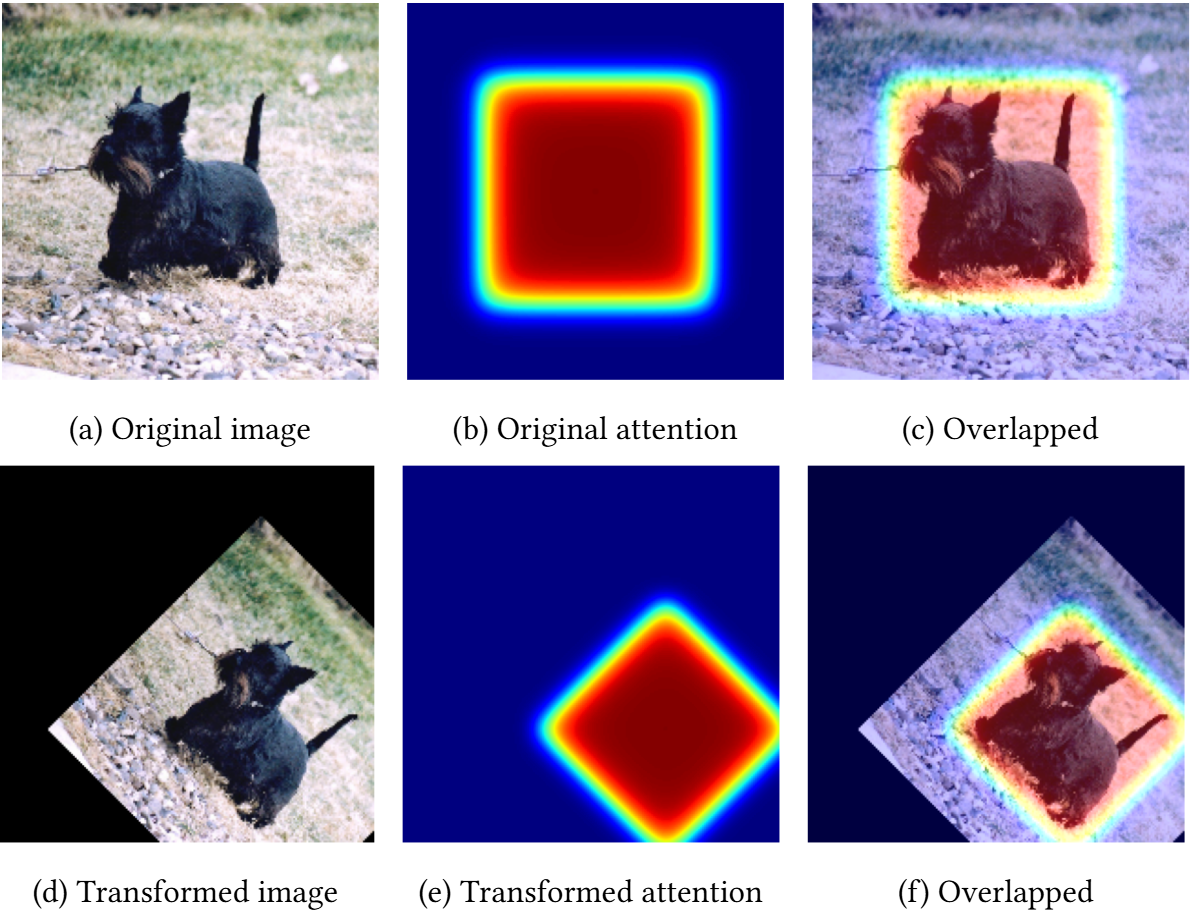}
\caption{Illustration of desirable equivariance property with respect to translation, rotation and scaling.}\label{fig:equivariance}
\end{figure}

When dealing with images, the attention map should respect some equivariance constraints. For example, if the input image is translated by ($\delta h, \delta w$), then the predicted attention map should undergo the same translation. Besides, this should hold true also for rotation and scaling. Indeed, when the input image is rotated by an angle $\delta \alpha$ or scaling by a factor, the support of attention map should be subjected to the same transformation. This is referred to as {\it equivariance property}. This is illustrated in Fig. \ref{fig:equivariance} (see also our qualitative results in Fig. \ref{fig:qualitative_eq} of Section \ref{sec:qualitative_eq}). 

Let us now discuss how we can constrain our model to respect this equivariance property.
In our method, the support of attention map is rectangular and parametrized by 5 parameters. These parameters can be easily constrained to respect the equivariance property. More precisely, let us consider an input image $\mathcal{I}$. The parameters outputted by the attention model is $(\mu,\sigma,\alpha)$. 
Then, we randomly perform transformations applied on $\mathcal{I}$ in the following order: rotation (around the image center), scaling and translation.
Denote the randomly transformed image by $\Tilde{\mathcal{I}}$.  The parameters outputted by the attention model for $\Tilde{\mathcal{I}}$ is denoted by $(\Tilde{\mu},\Tilde{\sigma},\Tilde{\alpha})$. Let $\delta\alpha\in(-\pi,\pi]$, $\delta_\sigma>0$ and $\delta \mu \in[-1,1]^2$ denote respectively the parameters defining the random rotation, the random scaling and the random translation. Notice that the components of $\delta \mu$ are scaled to range $[-1,1]$ and that the scaling factor $\delta_\sigma$ is the same for the two dimensions. We set further 
\begin{equation*}
   \hat{\mu}:= \begin{pmatrix} \hat{\mu}_1 \\ \hat{\mu}_2 \end{pmatrix} = \delta_\sigma\left(\mathcal{R}_{\delta \alpha}\begin{pmatrix} \mu_1 -c_1\\ \mu_2 -c_2\end{pmatrix}+\begin{pmatrix} c_1 \\  c_2 \end{pmatrix}\right) + \begin{pmatrix} \delta \mu_1 \\ \delta \mu_2 \end{pmatrix}, \\
\end{equation*}
\begin{equation*}
    \hat{\alpha}=\alpha + \delta \alpha,\;\;
    \hat{\sigma}:= \begin{pmatrix} \hat{\sigma}_1 \\ \hat{\sigma}_2 \end{pmatrix} = \begin{pmatrix} \delta_s \sigma_1 \\ \delta_s \sigma_2 \end{pmatrix} \ ,
\end{equation*}
where $\begin{pmatrix} c_1 \\  c_2 \end{pmatrix}$ is the image center (by default equal to $\begin{pmatrix} 0.5 \\  0.5 \end{pmatrix}$).



To constrain the model to satisfy almost the equivariance property, 
we may use the penalty function 
$\mathcal{L}_{\text{equivariance}}= \|(\Tilde{\mu},\Tilde{\sigma},\Tilde{\alpha})
- (\hat{\mu},\hat{\sigma},\hat{\alpha})\|^2 $. Notice that other discrepancy can be used rather that the squared Euclidean norm of the difference.
By randomly sampling different transformations for different batches during training, the model is trained to respect almost the equivariance property. Note that the main model is trained for a main task with a given loss function denoted by $\mathcal{L}_{\text{main}}$. $\mathcal{L}_{\text{equivariance}}$ is just used as an auxiliary loss. Hence, the global loss is $\mathcal{L} = \mathcal{L}_{\text{main}}+ \lambda \mathcal{L}_{\text{equivariance}}$ ($\lambda$ is a weighting parameter). By adding this auxiliary loss, we enforce the model to respect almost the equivariance property. This reduces significantly the solution space for the attention module and facilitates the optimization, allowing so a better attention.


Our equivariance constraints are primarily applied to reduced the solution space of the meta attention model. Interestingly, these equivariance constraints are related to self-supervised learning (SSL). SSL gains attention of researchers recently as it does not need annotation for training. This consists in training models on pretext tasks to have a rich representation of the data. Then, the pretrained model is used for a downstream task \cite{jing2020self, kolesnikov2019revisiting}. For example, a simple but effective pretext task is image rotations. By forcing the model to correctly predict the rotations, it learns the useful features for downstream tasks \cite{gidaris2018unsupervised}.  In our method, the attention model is integrated and unified with the main model. So, for the attention model to respect the equivariance property (w.r.t. translation, rotation and scaling), the main model is encouraged to learn and to contain this information in its features. This allows a better feature learning.

\subsection{Residual connection in our attention module: a safer choice}

In this section, we discuss briefly how residual connection in our attention can help learning, especially in the early stage of the training when we have a bad initialization, i.e., the attention map support does not cover a good discriminative part of image. Please refer to Appendix \ref{sec:res_discuss_app} for more details about the calculations as well as discussion.

Let $\Phi_{\theta}$ denote the part of the main model prior to the attention module, where $\theta$ is the set of parameters of $\Phi$. Recall that $\mathcal{L}_{\text{main}}$ is the loss function for the considered task. Let $x\in\mathbb{R}^{H\times W\times C}$ be an output of $\Phi$. Then $x$ is fed into the attention module to obtain $\hat{x}$ (see Eq. \eqref{eq:attention_function}). In the the following passage, we consider the case without rescaling step in Eq. \eqref{eq:rescale_attention}. However, all the arguments hold true by simply replacing $f$ by $\widetilde{f}$.

We can show that the partial derivative of the loss function $\mathcal{L}_{\text{main}}$ w.r.t $\theta_k\in\theta$ can be written as,
\begin{equation}\label{eq:attention_res}
\frac{\partial \mathcal{L}_{\text{main}}}{\partial \theta_k}
= \sum_{i,j,c} \frac{\partial \mathcal{L}_{\text{main}}}{\partial \hat{x}_{i,j,c}} \left(1+f(x)_{i,j}\right) \cdot \frac{\partial x_{i,j,c}}{\partial \theta_k}
+ \sum_{i,j,c} \frac{\partial \mathcal{L}_{\text{main}}}{\partial \hat{x}_{i,j,c}}\cdot x_{i,j,c}\cdot \frac{\partial f(x)_{i,j}}{\partial \theta_k} \ .    
\end{equation}

If we do not use the residual connection, we have that
\begin{equation}\label{eq:attention_no_res}
\frac{\partial \mathcal{L}_{\text{main}}}{\partial \theta_k}
= \sum_{c} \sum_{(i,j)\in \mathcal{S}} \frac{\partial \mathcal{L}_{\text{main}}}{\hat{x}_{i,j,c}} f(x)_{i,j} \cdot \frac{\partial x_{i,j,c}}{\partial \theta_k}
+ \sum_{i,j,c} \frac{\partial \mathcal{L}_{\text{main}}}{\hat{x}_{i,j,c}}\cdot x_{i,j,c}\cdot \frac{\partial f(x)_{i,j}}{\partial \theta_k}  \ ,    
\end{equation} 

\noindent
where $\mathcal{S}$ is the support of the attention map $f(x)$.

Let us consider the first term in Eq. \eqref{eq:attention_no_res}. Without residual connection, the model only uses information inside the support of the attention map to update $\theta$ ($(i,j)\in \mathcal{S}$). Hence, if we have a bad initialization, i.e., the support cover a non-informative part of the image, then we cannot use information outside the support. This makes the optimization more difficult. 

Now, we go back to our setup where we use the residual connection (gradient in Eq. \eqref{eq:attention_res}). Comparing the first term in Eq. \eqref{eq:attention_res} with Eq. \eqref{eq:attention_no_res}, we see that Eq. \eqref{eq:attention_res} contains a supplementary term, $\sum_{i,j,c} \frac{\partial \mathcal{L}}{\hat{x}_{i,j,c}} \cdot \frac{\partial x_{i,j,c}}{\partial \theta_k}$. This guarantees that we always have the information outside the support of the attention map to update $\theta$. This holds even with a bad initialization, especially in the early stage of the training. This justifies the use of the residual connection in our method.

Now, we analyze the second term in Eq. \eqref{eq:attention_res} and \eqref{eq:attention_no_res}. Notice that $f(x)$ is a function of $(\mu,\sigma,\alpha)$ (the parameters of the rectangular support). In turn, $(\mu,\sigma,\alpha)$ themselves are functions of $\theta$. Hence, the gradient w.r.t $\theta_k$ in the second term is very \textit{indirect} in the sense that $\theta$ is updated based on how the support is moved (represented by $(\mu,\sigma,\alpha)$). In the early stage of training, if we have a bad initialization, i.e., the attention map does not cover a good discriminative part of image, then the moving of the support provides very weak information to update $\theta$.

\section{Theoretical insights}\label{sec:theoretical}
\subsection{Generalization error of generating attention maps}\label{sec:error}
In this section, we derive theoretical insights for better understanding how well the attention module can generalize on new samples.
For this aim, let $(X,A)$ be the random element representing the input and its ground-truth attention map. Let further $S=\{(x_i,a_i)\}_{i=1}^{N}$ denote the sample consisting of $N>0$ independent copies of $(X,A)$.  Recall that we also have the predicted attention map $f(x)$ for each $x$.

Now, let $\Phi$ be some metric to measure the error of the predicted attention map $f(x)$ catching the ground-truth attention map $a$ ($\Phi$ will be properly defined later).
Furthermore, let $R(f) := \mathbb{E}[\Phi(f(X),A)]$ denote the generalization error and $\widehat{R}(f) := \frac{1}{N}\sum_{i=1}^{N}\Phi(f(x_i),a_i)$ denote the empirical error on the sample $S$. This allows us to formally introduce the main objective of this section as follows.

\noindent
\textbf{Objective.} {\it We seek to quantify the upper bound of $R(f)$, depending on $\widehat{R}(f)$ and the functional set that produces the attention maps.}

\noindent
This shall be provided in Theorem \ref{theo:missing_rate}.

Notice that we do not have an explicit ground-truth attention map $a$ associated to each $x$. However, we assume that
\begin{equation}\label{assumption:h1}
\text{(H1)} \; \; \text{A small } \mathcal{L}_{\text{main}}  \text{ on training set implies } \text{a small } \widehat{R}(f) \ .    
\end{equation}
We recall that $\mathcal{L}_{\text{main}}$ is the loss function for the main task.

This assumption means that, if the model is well trained to have small training error in the final predictions, then the error $\Phi$ of attention maps (in some sense) is also small, at least on the training set. 
This is a reasonable assumption, as the model needs to pay attention to the important parts of the input to produce a good prediction for the main task.

Next, we shall define a proper $\Phi$ and other relevant definitions to properly quantify $R(f)$. For the sake of simplification, we assume that all the inputs are of spatial dimension $H\times W$.  Nevertheless, the  arguments remain true for arbitrary input spatial dimensions. We recall that the attention value for each pixel in the input images is in the range $[0,1]$. However, in practice, the predicted attention value for each position in the image is very close either to $0$ (no attention for unimportant parts), or to $1$ (high attention for important parts). Thus, for simplification, we assume that the generated attention map is  binary valued in $\{0,1\}$. Equivalently, for the appropriate definitions that follow, we define the binary value of attention as $\{-1,1\}$ instead of $\{0,1\}$. The value of $1$ corresponds to attention parts and $-1$ otherwise. We define ground-truth attention values in the same way. That is, for an input image of spatial size $H\times W$, both the generated and ground-truth attention maps are in $\{-1,1\}^{H\times W}$. Now, to measure how well the generated attention {\it catch up} the ground-truth one, we define the \textit{catching rate function} as follows.
\begin{definition}[Catching rate function] \label{def:catch}
    For any fixed $H,W\in \mathbb{N}^+$, let $\Omega = \{-1,1\}^{H\times W}$. We define the catching rate function $\Psi:  \Omega \times \Omega \mapsto \mathbb{R}$ for any $ a,\Tilde{a} \in \Omega$ as
    \begin{equation}\label{eq:catching_rate}
        \Psi(a,\Tilde{a}) = \frac{1}{H\times W} \sum_{i=1}^{H} \sum_{j=1}^{W} a_{i,j}\times \Tilde{a}_{i,j}\ .
    \end{equation}
\end{definition}

\begin{remark} \label{remark:catch}
We remark that,
\begin{itemize}
    \item $\Psi$ is symmetric ($\Psi(a,\Tilde{a})=\Psi(\Tilde{a},a)$) and $-1 \leq \Psi(a,\Tilde{a}) \leq 1, \ \forall a,\Tilde{a} \in \Omega$.
    \item If we consider $a$ as the ground-truth attention mask and $\Tilde{a}$ as the predicted one, then $\Psi(a,\Tilde{a})=1$ iff $a=\Tilde{a}$, i.e., the prediction perfectly catch the ground-truth one. Likewise, $\Psi(a,\Tilde{a})=-1$ iff $a=-\Tilde{a}$, corresponding to the case of total missing. That is, ground-truth important parts ($a_{i,j}=1$) is predicted as unimportant ($\Tilde{a}_{i,j}=-1$) and inversely for the ground-truth unimportant parts.
    \item If $-1 < \Psi(a,\Tilde{a}) < 1$, the predicted attention map is partly correct. The higher the value of $\Psi$, the more the predicted attention map catches up with the ground-truth one.
\end{itemize}
\end{remark}

In order to have a missing rate indicator valued in $[0,1]$ we set
\begin{equation}\label{eq:missing_rate}
        \Phi(a,\Tilde{a}) = \frac{1}{2} \left(1 - \Psi(a,\Tilde{a}) \right) \ .
    \end{equation}
Indeed, we have that  this last function is equal to $0$ in  case of no missing (total catch-up) and to $1$ in  case of total missing. 

In order to formalize our theoretical results, let introduce some notations.
For $\mathcal{A}_{\text{in}} \subseteq \Omega$, let $(X,A) \in \mathcal{X}\times \mathcal{A}_{\text{in}}$ be the random element representing the input and its attention map.
For an $\mathcal{A} \subseteq \Omega$, let us define $\mathcal{F}_{\mathcal{A}}$ as the set of all the functions on $\mathcal{X}$ with values in $\mathcal{A}$. In equation,
\begin{equation}
\mathcal{F}_{\mathcal{A}} = \{f: f(\mathcal{X}) \subseteq \mathcal{A} \subseteq \Omega\} \ .    
\end{equation}
Let  $\mathcal{A}_{\text{ours}}$ denote the set of all attention maps (of spatial dimension $H \times W$) with values $1$ in some rectangle and $-1$ outside this rectangle (with the binary simplification). Then, for our method, the involved functional set consists of all the functions valued in $\mathcal{A}_{\text{ours}}$.  In the case of position-wise method, $\mathcal{A}_{\text{pw}}$ is simply $\Omega$.

\begin{definition}[Relevance level]
For $(\mathcal{A},\mathcal{A}_{\text{in}}) \subseteq \Omega^2$, we define the relevance level of $\mathcal{F}_{\mathcal{A}}$ w.r.t. $\mathcal{X}\times \mathcal{A}_{\text{in}}$ as,
    \begin{equation}\label{eq:relevance_level}
        \rho_{\mathcal{F}_{\mathcal{A}}} = \inf_{f\in\mathcal{F}_{\mathcal{A}}, (x,a)\in \mathcal{X}\times \mathcal{A}_{\text{in}}} \Psi(f(x),a) \ .
    \end{equation}
\end{definition}
\noindent
Obviously,
\begin{equation}\label{cor:relevance}
\text{if } \mathcal{A}_1\subseteq \mathcal{A}_2, \text{then } \rho_{\mathcal{F}_{\mathcal{A}_1}} \geq \rho_{\mathcal{F}_{\mathcal{A}_2}}   \ .
\end{equation}
Indeed, if  $\mathcal{A}_1\subseteq \mathcal{A}_2$, then $\mathcal{F}_{\mathcal{A}_1} \subseteq \mathcal{F}_{\mathcal{A}_2}$. Moreover, from Eq. \eqref{eq:relevance_level}, we remark that if $\mathcal{F}_{\mathcal{A}_1} \subseteq \mathcal{F}_{\mathcal{A}_2}$, then $\rho_{\mathcal{F}_{\mathcal{A}_1}} \geq \rho_{\mathcal{F}_{\mathcal{A}_2}}$. Hence, we have the inequality in Eq. \eqref{cor:relevance}.

\noindent
The next theorem relates the expected missing rate to the empirical one obtained on the training sample and the relevance level.


\begin{theorem} [Generalization missing rate] \label{theo:missing_rate} 
For any $\delta > 0$, with probability at least $1- \delta$, the following holds for all $f \in \mathcal{F}_{\mathcal{A}}$:
\begin{equation}\label{eq:upper_bound}
R(f) \leq \widehat{R}(f) + \frac{1-\rho_{\mathcal{F}_\mathcal{A}}}{2} +3\sqrt{\frac{\log\frac{2}{\delta}}{2N}} \ .
\end{equation}
\end{theorem}

\begin{proof}
See Appendix \ref{sec:proof_error}.
\end{proof}

\begin{remark}
\ 
\begin{itemize}
    \item From the second term in the RHS of Eq.\eqref{eq:upper_bound}, we see that the upper bound of the generalization error (expected missing rate) of a family $\mathcal{F}$ is a decreasing function of its relevance level. That is, the higher its relevance level, the lower its generalization error upper bound.
    \item If we have some prior knowledge about $\mathcal{A}_{\text{in}}$, we can constrain $\mathcal{A}$ to increase the relevance level of $\mathcal{F}_\mathcal{A}$. This allows for decreasing the generalization error upper bound, hence better generalization capacity.
\end{itemize} 
\end{remark}

From the statement \eqref{cor:relevance}, we see that a straightforward way to increase the relevance level of a family $\mathcal{F}_{\mathcal{A}}$ is to reduce its output space $\mathcal{A}$. In the case of position-wise method, if there is no particular constraint, $\mathcal{A}_{\text{pw}}=\Omega$ (\textit{pw} stands for \textit{position-wise}). That is, each pixel can be affected attention value of $1$ or $-1$. This makes the relevance level low, thus difficult to generalize to new examples. In contrast, with our method, the attention map is constrained to have a rectangular support. This reduces drastically the output space $\mathcal{A}$, thus increasing the relevance level.

Now, a natural question follows: \textit{Why don't we simply reduce the output space $\mathcal{A}$, for example, to a singleton to increase the relevance level?} In fact, beside the relevance level, the flexibility of a hypothesis family is also important. In the singleton case, the predicted attention map is constant for all input $x$. Hence, it can not adapt to each input. Consequently, the empirical error $\widehat{R}(f)$ is high. This makes the upper bound worse. Hence, the balance between flexibility and relevance level is important. With our method, we assume that the attention support is rectangular, and we allow it to freely move around (in the size of the rectangle as well as in its position). This allows for sufficient flexibility but also for a reasonable relevance level.
Moreover, if we have some prior knowledge about $\mathcal{A}_{\text{in}}$, we can further reduce the output space $\mathcal{A}_{\text{our}}$ without sacrifying the flexibility. For example, if we know that the interested object is always in certain zones of the input images, we can constrain the rectangle to move and to orientate itself correctly in these zones. This increases the level of relevance. At the same time, flexibility is ensured because we allow the rectangle to have sufficient capacity to cover these interested areas. In contrast, for position-wise method, the attention is computed separately for each position, so handling the predicted attention zone is not trivial.

\subsection{A simplified probabilistic attention model in the context of classification}

In the last section, we make the assumption \eqref{assumption:h1} which assumes that, to have a small error in the final predictions, the predicted attention map should well \textit{catch up} the ground-truth one. In this section, we propose a simplified  probabilistic framework to support this intuition. More precisely, in the context of classification, we shall show that if the predicted attention maps get closer to the ground-truth maps, the conditional distributions of different classes get more separated from each other, which facilitates the final classification. Hence, it is expected that when the final classification loss is minimized, the predicted maps should move closer to the ground-truth map, as this is the direction which facilitates the final classification. This gives a solid support to the assumption in Eq. \eqref{assumption:h1}. 

Now, for the sake of simplicity, we suppose that each $x$ is associated with one ground-truth deterministic attention map $a(x)$, i.e., $\mathbb{P}(A=a(x)|X=x)=1$. That is, knowing $x$, we can define its associated attention map using an implicit function $a(\cdot)$. Recall that for each $x$, we also have at hand a predicted attention map $f(x)$. Now, let $\mathcal{S}(x)$ be the support of $a(x)\in \{-1,+1\}^{H\times W}$, i.e., $\mathcal{S}(x)=\{(i,j)\in[H]\times[W]: a(x)_{i,j}=1\}$. Analogously, let $\widehat{\mathcal{S}}(x)$ be the support of the predicted map $f(x)$. For the sake of brevity, we denote the attention model in the functional form as $\mathcal{M(\cdot)}$. That is, 
\begin{equation}\label{eq:functional_form}
\widehat{x}=\mathcal{M}(x) \ .
\end{equation}
We further assume that $\mathcal{M}$ is injective so that the preimage $\mathcal{M}^{-1}(\widehat{x})$ is a singleton. For the sake of brevity, please refer to Appendix \ref{sec:proof_injective} where we provide generic conditions such that this assumption holds true. We now can write $\mathcal{S}(x)=\mathcal{S}(\mathcal{M}^{-1}(\widehat{x}))$ and the same for $\widehat{\mathcal{S}}$. Moreover, let $\widehat{\mathcal{X}}$ denote the space generated from $\mathcal{X}$ by $\mathcal{M}$. That is,  $\widehat{\mathcal{X}}:=\{\mathcal{M}(x):x\in\mathcal{X}\}$.

Notice that each image can be decomposed into two main parts: the semantic part and the background.  The semantic part contains discriminative features. On the contrary, the background is common for all the classes and does not contain relevant information to distinguish one class from another. Thus, if the predicted attention correctly captures the important elements of the inputs, then the features of different classes should be more separated from each other. Otherwise, if the predicted map fails to catch up with the ground-truth map, the background information should come into play and {\it contaminates} the discriminative information. Placing ourselves in the output space $\widehat{\mathcal{X}}$, we propose to model the conditional density (w.r.t. to a reference measure $\nu$) of the output, given a class $y$ as follows,

\begin{equation}\label{eq:mixture_model}
p_y(\widehat{x}) := p(\widehat{x}|Y=y)
= \lambda(\widehat{x})\cdot p^{\text{pure}}_y(\widehat{x}) + \eta_y\left(1-\lambda(\widehat{x})\right)\cdot p_{\text{background}}(\widehat{x}) \ ,    
\end{equation}

\noindent
where
\begin{eqnarray}
\lambda(\widehat{x}) = \frac{|\mathcal{S}(\mathcal{M}^{-1}(\widehat{x}))\cap \widehat{\mathcal{S}}(\mathcal{M}^{-1}(\widehat{x}))|}{|\mathcal{S}(\mathcal{M}^{-1}(\widehat{x}))\cup \widehat{\mathcal{S}}(\mathcal{M}^{-1}(\widehat{x}))|} \ .
\end{eqnarray}

\noindent
Here, $|\mathcal{S}|$ denotes the cardinality of a set $\mathcal{S}$ (number of elements in $\mathcal{S}$).  $\eta_y$ is a normalization factor (such that $\int p_y(\widehat{x})d\nu(\widehat{x})=1$). So that,
\begin{eqnarray*}\label{eq:norm_factor}
\eta_y 
= \frac{1-\int_{\widehat{\mathcal{X}}} \lambda(\widehat{x})p^{\text{pure}}_y(\widehat{x})d\nu(\widehat{x})}{1-\int_{\widehat{\mathcal{X}}}\lambda(\widehat{x})p_{\text{background}}(\widehat{x})d\nu(\widehat{x})} \ ,
\end{eqnarray*}
that is assumed to be defined. Notice also that $p_{\text{background}}$ is common for all the classes, contrarily to $p^{\text{pure}}_y$. For the sake of brevity, we use $x$ instead of $\mathcal{M}^{-1}(\widehat{x})$ in some following passages.

\noindent
\textbf{Interpretation.} Our modeling has the following desirable properties.
\begin{enumerate}
    \item When $f(x)$ perfectly catches up $a(x)$, we have $\mathcal{S}(x)= \widehat{\mathcal{S}}(x)$. Hence, $p_y=p^{\text{pure}}_y$. No background information contaminates the pure conditional distribution (as $1-\frac{|\mathcal{S}(x)\cap \widehat{\mathcal{S}}(x)|}{|\mathcal{S}(x)\cup \widehat{\mathcal{S}}(x)|}=0$).
    \item When $f(x)$ does not perfectly catch up $a(x)$, we have $1-\frac{|\mathcal{S}(x)\cap \widehat{\mathcal{S}}(x)|}{|\mathcal{S}(x)\cup \widehat{\mathcal{S}}(x)|}>0$. In this case, some background information contaminates the pure distribution.
    \item When $f(x)$ totally fails to catch up $a(x)$, $|\mathcal{S}(x)\cap \widehat{\mathcal{S}}(x)|=0$. Thus, no semantic information is captured and we get only the background information.
    \item If $\mathcal{S}(x) \subseteq \widehat{\mathcal{S}}(x)$ but $|\mathcal{S}(x)|<<|\widehat{\mathcal{S}}(x)|$, $\frac{|\mathcal{S}(x)\cap \widehat{\mathcal{S}}(x)|}{|\mathcal{S}(x)\cup \widehat{\mathcal{S}}(x)|}$ is small. This corresponds to the case where the predicted support contains the true support, but the predicted support also covers a large part of the background. So, the background information contaminates the pure distribution.
\end{enumerate}

\noindent
This justifies the use of this modeling.

\noindent
In the next proposition, we relate $\frac{|\mathcal{S}(x)\cap \widehat{\mathcal{S}}(x)|}{|\mathcal{S}(x)\cup \widehat{\mathcal{S}}(x)|}$ to $f(x)$ and $a(x)$.


\begin{prop}\label{prop:union_intersection}
We have
\begin{equation}
\frac{|\mathcal{S}(x)\cap \widehat{\mathcal{S}}(x)|}{|\mathcal{S}(x)\cup \widehat{\mathcal{S}}(x)|} = \frac{\Psi((a(x)+\mathbf{1})/2,(f(x)+\mathbf{1})/2)}{2\Phi((\mathbf{1}-a(x))/2,(\mathbf{1}-f(x))/2)} \ .     
\end{equation}
Here, $\mathbf{1}$ is the matrix having the same size as $a(x)$ (and so $f(x)$) whose all entries equal to $1$. That is, $(a(x)+\mathbf{1})_{i,j}= a(x)_{i,j}+1$.     
\end{prop}

\begin{proof}
    See Appendix \ref{sec:proof_union_intersection}.
\end{proof}

This leads to the following definition.
\begin{definition}[Fitting rate]
Let $a_1,a_2\in\{-1,+1\}^{H\times W}$, the fitting rate between $a_1$ and $a_2$ is defined as
\begin{equation}
\widetilde{\Psi}\left(a_1,a_2\right)=\frac{\Psi((a_1+\mathbf{1})/2,(a_2+\mathbf{1})/2)}{2\Phi((\mathbf{1}-a_1/2,(\mathbf{1}-a_2/2)} \ . 
\end{equation}
\end{definition}

Obviously, $0\leq\widetilde{\Psi}\left(a_1,a_2\right)\leq1$. Moreover, $\widetilde{\Psi}\left(a_1,a_2\right)=1$ iff $a_1=a_2$ (that is, they share the same support, i.e., $\mathcal{S}_{a_1}= \widehat{\mathcal{S}}_{a_2}$). $\widetilde{\Psi}\left(a_1,a_2\right)$ gets larger as $a_2$ gets more similar to $a_1$.

We now provide theoretical insights about the conditional distribution of each class. Let $\mathcal{D}_y$, $\mathcal{D}_y^{\text{pure}}$ and $\mathcal{D}_{\text{background}}$ denote the distributions associated to the density functions $p_y$, $p_y^{\text{pure}}$ and $p_{\text{background}}$, respectively. From a probabilistic viewpoint, we shall show that the conditional distributions $\mathcal{D}_y$'s become more distinct from each other as predicted attention maps approach the ground-truth ones. To measure the divergence between distributions, we use the total variation distance defined as follows.

\begin{definition}[Total variation distance]
Let $\mathcal{D}_1$ and $\mathcal{D}_2$ be probability measures on a measurable space $(\mathcal{X},\sigma)$. The total variation (TV) distance between  $\mathcal{D}_1$ and $\mathcal{D}_2$ is defined as
$$
    d_{TV}(\mathcal{D}_1, \mathcal{D}_2) = \sup_{\Gamma\in \sigma} |\mathcal{D}_1(\Gamma)-\mathcal{D}_2(\Gamma)| \ .
$$
\end{definition}

\noindent
Here, $\sigma$ is the $\sigma-$algebra where all the probability distributions are defined.

The following theorem relates the TV distance between the conditional distributions to the fitting rate between predicted and true attention maps.
\begin{theorem}\label{theo:divergence_attention} 
Consider a pair of classes $(y,y')$.
\begin{enumerate}
\item We have
\begin{equation*}
d_{TV}(\mathcal{D}_y,\mathcal{D}_{y'}) \geq \\ \inf_{x\in\mathcal{X}}\widetilde{\Psi}(a(x),f(x)) \cdot d_{TV}(\mathcal{D}_y^{\text{pure}},\mathcal{D}_{y'}^{\text{pure}}) .  
\end{equation*}
\item If $\mathcal{D}_y^{\text{pure}}$ and $\mathcal{D}_{y'}^{\text{pure}}$ have distinct supports, then
\begin{equation*}
d_{TV}(\mathcal{D}_y,\mathcal{D}_{y'}) \geq
\min_{\widehat{y}\in\{y,y'\}} \left( \mathbb{E}_{\widehat{x}\sim \mathcal{D}_{\widehat{y}}^{\text{pure}}}\left[\widetilde{\Psi}(a(\mathcal{M}^{-1}(\widehat{x})),
f(\mathcal{M}^{-1}(\widehat{x})))\right] \right)  .
\end{equation*}
    
\end{enumerate}
\end{theorem}

\begin{proof}
See Appendix \ref{sec:proof_divergence_attention}.
\end{proof}

\begin{remark} \
\begin{itemize}
\item The statement (1) in Theorem \ref{theo:divergence_attention} provides a nice decomposition for lower bound on the TV distance between the conditional distributions. On the one hand, it depends on how well the predicted maps fit to the ground-truth maps (represented by $\inf_{x\in\mathcal{X}}\widetilde{\Psi}(a(x),f(x))$). On the other hand, it also depends on the ideal TV distance ($d_{TV}(\mathcal{D}_y^{\text{pure}},\mathcal{D}_{y'}^{\text{pure}})$). We get this ideal distance when we have only semantic information, and no background is contaminating the distribution.
\item Indeed, if we have perfect predicted attention maps, but the ideal pure conditional distributions themselves are not well separated from each other, then we still cannot separate the classes.
\item In the statement (2) of Theorem \ref{theo:divergence_attention}, it is assumed that the pure conditional distributions have distinct supports. While this might seem to be a strong assumption, it is usual in the context of deep learning. Indeed, we work in the feature space. Hence, the model is trained to well separate the features of different classes from each other.
\item If this last condition is fulfilled, then the result of the statement (2) is stronger than that of the statement (1), as we get the expectation of the fitting rate instead of its infimum over the space.
\item Notice that in the binary classification where the two classes $y$ and $y'$ are equiprobable, the optimal classification accuracy is equal to $(1+d_{TV}(\mathcal{D}_y,\mathcal{D}_{y'}))/2$ (see for example \cite{pablo}). Hence, the optimal classification accuracy increases as the predicted attention maps approach the ground-truth ones.
\end{itemize}
    
\end{remark}

\section{Experiments}\label{sec:exp}

In this section, we perform experiments on the Oxford-IIIT Pet dataset \cite{parkhi2012cats}.  Oxford-IIIT Pet contains 37 classes. These are different breeds of cat and dog.
This dataset contains images of different sizes (3 channels). All images are resized to $300\times300$ pixels for training and testing. In total, we have 7349 images divided in 37 classes. We intentionally choose this dataset because the captured animals in each image have a large variations in scale, pose and lighting, as explained by the authors. Hence, this can demonstrate the flexibility of our method for capturing objects in different scales and poses.

For training, we use the cross-entropy loss for the classification task, along with the equivariance loss. That is, $\mathcal{L} = \mathcal{L}_{\text{main}} + \lambda\mathcal{L}_{\text{equivariance}}$, where $\mathcal{L}_{\text{main}}$ is the standard cross-entropy loss. We set $\lambda=0.1$ in this experiment. We also set $\lambda=0$ (no equivariance constraint) to see the impact of these equivariances. For the scaling scale  (in Eq. \eqref{eq:final_function}) for generating attention maps, we fixed $s=6$.
We test on two different models EfficientNet-b0 \cite{tan2019efficientnet} and MobileNetV3 \cite{howard2019searching}. EfficientNet-b0 is approximately twice larger than MobileNetV3 in terms of running time. Thus, by testing on different models with different sizes and structures, we get a better view on the performance of our method.

We follow \cite{park2018bam} for the design of position-wise module. To demonstrate that our method can learn well with fewer parameters, we use a design with lower complexity compared to the position-wise module.  All the details for conducting our experiments can be found in Appendix \ref{sec:exp_details}.

\subsection{Quantitative Evaluation}

We divide the dataset into training/validation set with different ratios: 6:4, 7:3, 8:2. For each of the ratio, we perform 5 independent runs (i.e. 5 different splits for each ratio). For each run, we choose the corresponding ratio for each class for training/validation (to make sure that each class in training and validation set represents well the number of examples in each class). We test 4 methods: the base model without attention module, the base model with position-wise attention module, the base model with our rectangular attention where we test with and without the equivariance constraints. The results are shown on MobiletNetV3 and Efficient-b0 in Table \ref{tab:oxford_pets}. 

\begin{table*}[ht]
\caption{Validation accuracy at different train/val ratio on Oxford-IIIT Pet dataset using MobilenetV3 and EfficientNet-b0. \textbf{EQVC}: equivariance constraints.} \label{tab:oxford_pets}
\centering
\resizebox{1.0\textwidth}{!}
{
\begin{tabular}{|c|c|c|c|c|}
\hline
Model & train/val ratio & $6:4$ & $7:3$ & $8:2$ \\ \hline
\multirow{4}{*}{MobileNetV3} & no attention &$91.153 \pm 0.305$ & $92.344 \pm 0.442$ & $92.630 \pm 0.515$ \\ \cline{2-5}
& position-wise attention & $91.525 \pm 0.170$ & $92.218 \pm 0.345$ & $92.981 \pm 0.561$ \\ \cline{2-5}
& rectangle attention - no EQVC (ours) & $91.755 \pm 0.201$ & $92.678 \pm 0.198$ & $93.185 \pm 0.632$  \\ \cline{2-5}
& rectangle attention - with EQVC (ours) & $\mathbf{91.823} \pm 0.335$ & $\mathbf{92.822} \pm 0.277$ & $\mathbf{93.225} \pm 0.620$ \\ \hline
\multirow{4}{*}{EfficientNet-b0} & no attention   & $93.459 \pm 0.229$ & $94.085 \pm 0.364$ & $94.577 \pm 0.327$  \\ \cline{2-5}
& position-wise attention & $93.290 \pm 0.320$ & $93.922 \pm 0.567$ & $94.455 \pm 0.245$ \\ \cline{2-5}
& rectangle attention - no EQVC (ours) & $93.811 \pm 0.319$ & $94.355 \pm 0.396$ & $94.739 \pm 0.406$  \\ \cline{2-5}
& rectangle attention - with EQVC (ours) & $\mathbf{93.872} \pm 0.414$ & $\mathbf{94.355} \pm 0.422$ & $\mathbf{94.915} \pm 0.493$ \\ \hline
\end{tabular}
}
\end{table*}

From Table \ref{tab:oxford_pets}, a first remark is that our method systematically outperforms the case with no attention module or with position-wise attention module. Moreover, equivariance constraints seem to improve the performance of our method. On MobileNetV3, in the cases of split ratio 6:4 and 8:2, the position-wise method seems to improve slightly the performance of the model. However, in the case of train/validation ratio 7:3, adding position-wise spatial attention module slightly decreases the performance of the model, compared to the case without attention module. On EfficientNet-b0 which is twice larger than MobileNetV3, adding position-wise attention decreases the model performance, compared to not using the attention module (see Table \ref{tab:oxford_pets}). This remark is interesting, implies that on larger model, the position-wise attention is not necessarily useful for improve the model performance. This reinforces our conjecture that the position-wise method tends to generate attention map with irregular boundaries (that will be shown in the next section). Hence, the module struggles with new examples. In contrast, on EfficientNet-b0, our method still improves the model performance. This strongly suggests the stability of our method.

\begin{figure}[ht]
\centering
\includegraphics[width=\textwidth]{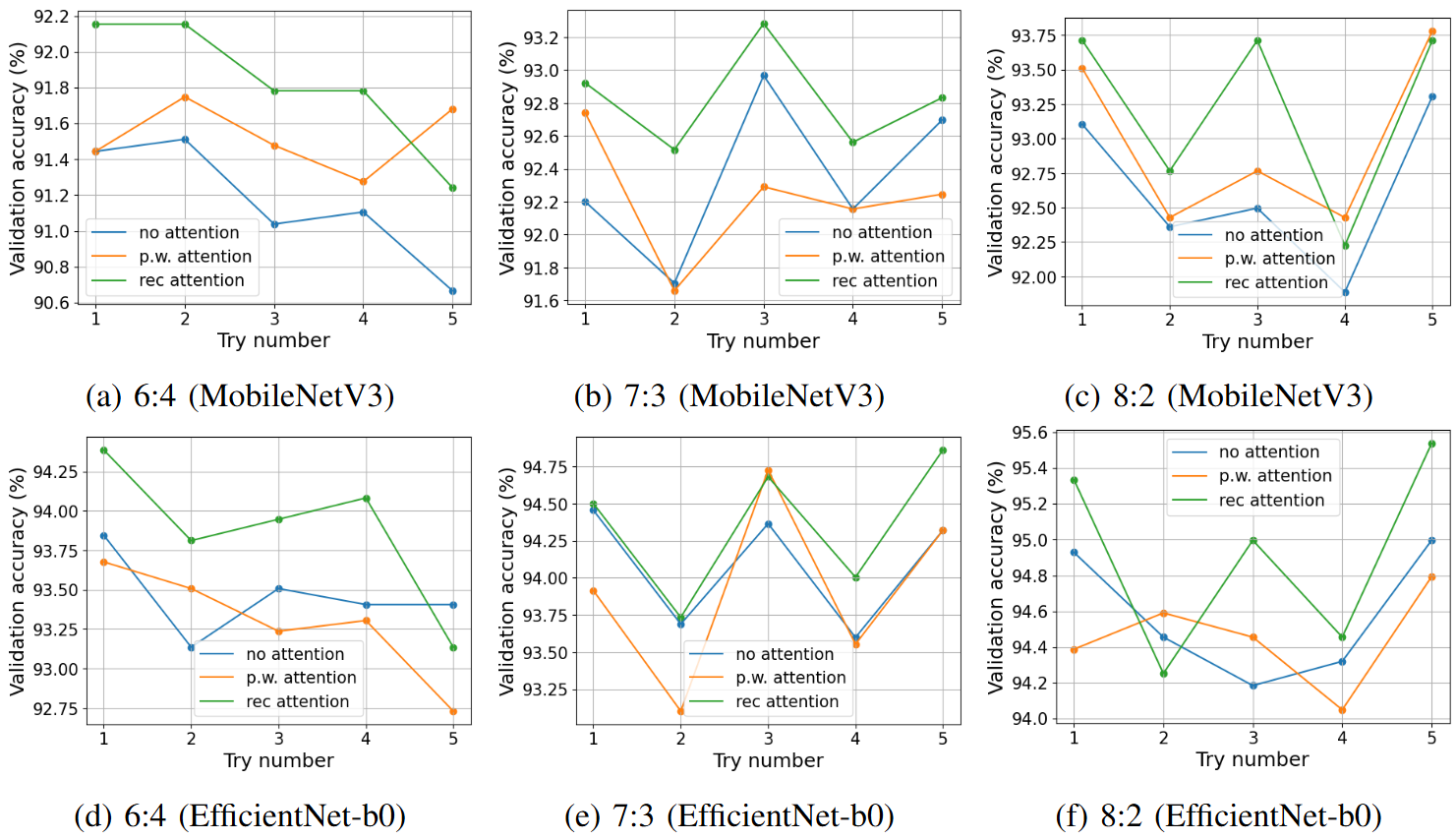}

\caption{Validation accuracy  on Oxford-IIIT Pet dataset for different train/validation split ratios (6:4; 7:3 and 8:2). For each split ratio, we perform 5 independent runs using 5 different random train/validation splits. \textbf{p.w. attention} and \textbf{rec attention} stand for position-wise attention and rectangular attention, respectively.}\label{fig:compare_acc_oxford_pets}
\end{figure}

Moreover, for each train/validation split ratio, the validation accuracies over 5 splits are plotted in Fig. \ref{fig:compare_acc_oxford_pets}. We remark that our method outperforms the two other ones in the majority of cases, implying its superiority. We also remark that the validation accuracy variances is due to the fact that there are some train/val splits more difficult than others, which are creating struggles for the 3 methods.

\subsection{Qualitative Evaluation}

\begin{figure}[ht]
\centering
\includegraphics[width=0.9\textwidth]{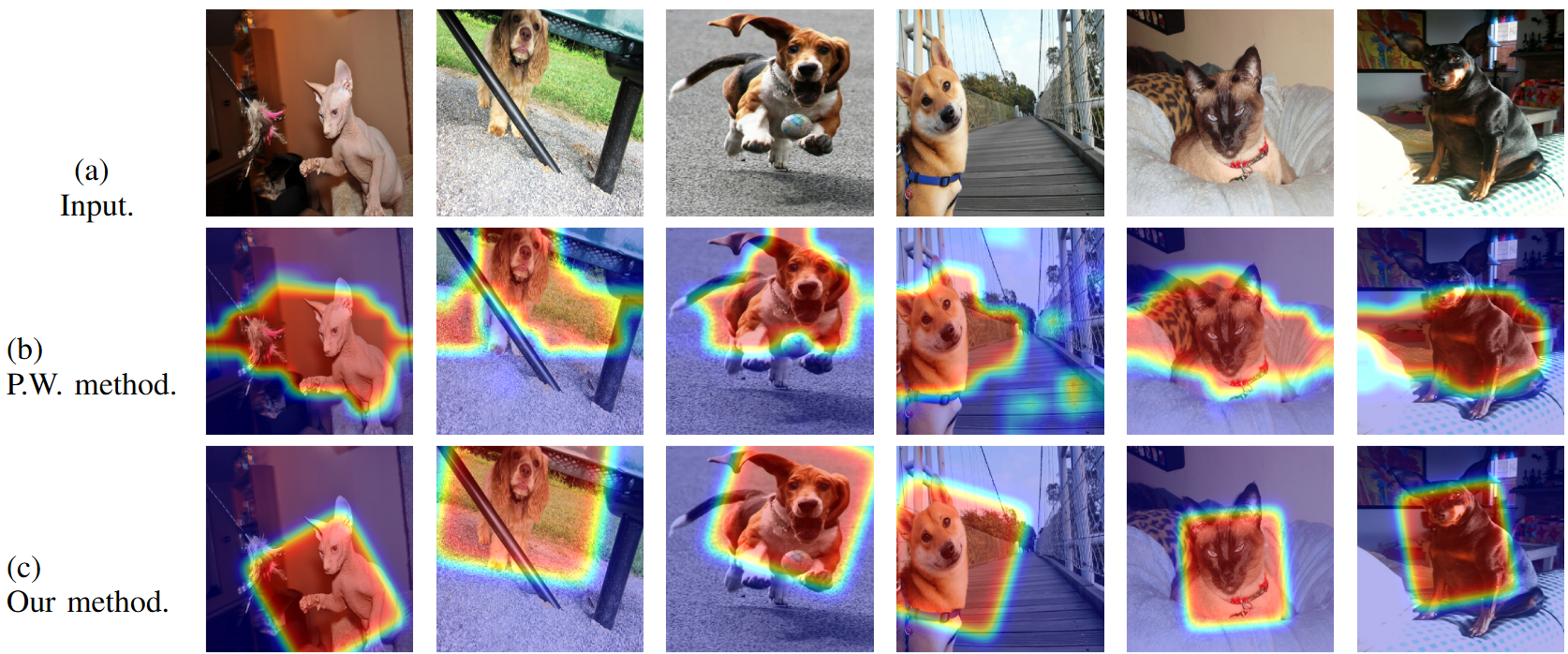}
\caption{Qualitative evaluation of attention maps generated by position-wise (P.W.) method and our method on validation set (Oxford-IIIT Pets). From top to bottom: input, P.W. method, our method.}\label{fig:qualitative_oxford_pets}
\end{figure}

We now evaluate qualitatively the attention maps generated by our method and position-wise method (applied on MobileNetV3). For this objective, we superimpose the attention map of each method on the input image.
Results for randomly chosen images of validation set are shown in Fig. \ref{fig:qualitative_oxford_pets}.

In Fig. \ref{fig:qualitative_oxford_pets}, a first remark is that the position-wise method tends to generate irregular attention maps. It can also produce some false positives (for example, the $1^{st}$ and $4^{th}$ columns of Fig.  \ref{fig:qualitative_oxford_pets}), where some background has high attention. This is not surprising, as the attention value for each position is generated separately. Hence, one has no guarantee about the continuity and regularity of the resulted attention map. This may explain why this module make the model performance worse in some cases in the above quantitative experiment. In contrast, our method generates attention maps with nearly-rectangular support by construction. This allows a more stable boundary. We see that the support adapts very well to capture the discriminative part in images.

\subsection{Impact of the layer depth on attention maps}
\begin{figure}[ht]
\centering
\includegraphics[width=0.75\textwidth]{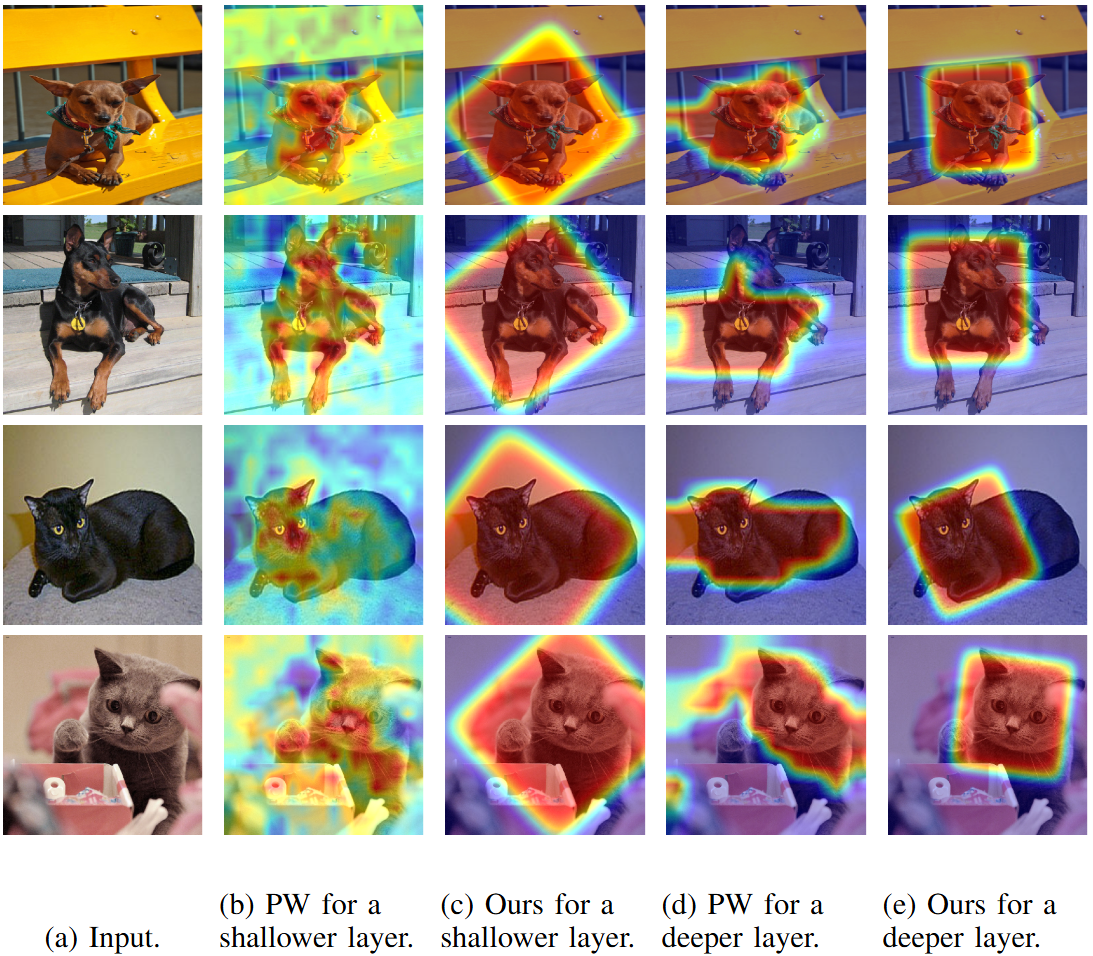}
\caption{Comparing impact of layer depth on attention maps computed for validation set. \textbf{First column:} input image, \textbf{second and third column:} position-wise and our rectangular attention for a shallower layer, \textbf{forth and fifth column:} position-wise and our rectangular attention for a deeper layer.}\label{fig:compare_layer}
\end{figure}

Our spatial attention module can be integrated after any convolutional layer. Obviously, in order for the module to locate correctly the interested object for a final correct prediction, this module should be integrated in a sufficiently deep layer. As such, the features contain sufficiently rich semantic information to provide to the attention module. Otherwise, it could be difficult for the module to find a correct region. Now, we experiment by integrating the attention module in a shallower layer. We compare our method with the position-wise counterpart. The results are shown in Fig. \ref{fig:compare_layer}. For the deeper layer (forth and fifth columns), the two method seems to capture more or less well the objects. Obviously, the position-wise method provides irregular attention maps, as explained previously. Now let us observe the results for the shallower layer (the  second and third columns for the position-wise and our methods, respectively). An interesting remark is that this layer allows us to see a remarkable difference between the two methods. With the position-wise method, it provides attention maps that are very discontinue and fails to focus on the interesting regions. This is because the features of this shallower layer are not semantically informative enough. However, our method still succeeds to locate the interesting objects in images. This is because the attention region of our method is sparingly characterized by only 5 parameters. Moreover, the attention region is constrained to be in a continuous rectangular shape. This makes the localization of attention maps easier and more stable. For the position-wise method, when the features are not semantically presented, it is difficult to generate a continuous and stable map.

This experiment strongly suggests that our method provides more stable results than the standard position-wise method. Moreover, with the same shallow layer, our method correctly locates the interesting region, whereas the position-wise counterpart completely fails. This also suggests that the two methods have completely different learning mechanisms.

\subsection{Qualitative evaluation of equivariance property}\label{sec:qualitative_eq}
\begin{figure*}[ht]
\centering
\begin{subfigure}{\textwidth}
    \centering
    \parbox[r][-0.1\textwidth]{.1\textwidth}{\caption{\\}\label{fig:quali_eq1}}
    \includegraphics[width=0.7\textwidth]{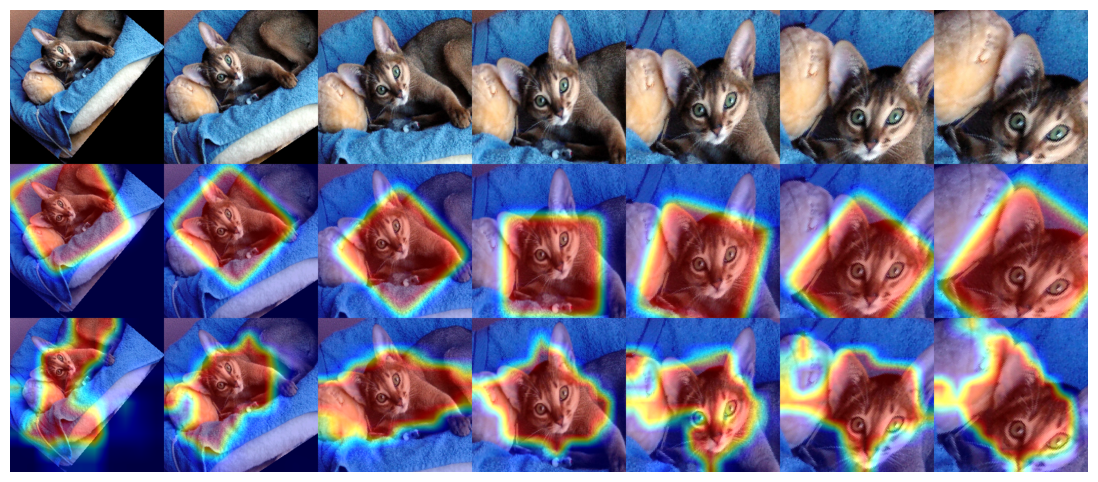}
\end{subfigure}
\begin{subfigure}{\textwidth}
    \centering
    \parbox[r][-0.1\textwidth]{.1\textwidth}{\caption{\\}\label{fig:quali_eq2}}
    \includegraphics[width=0.7\textwidth]{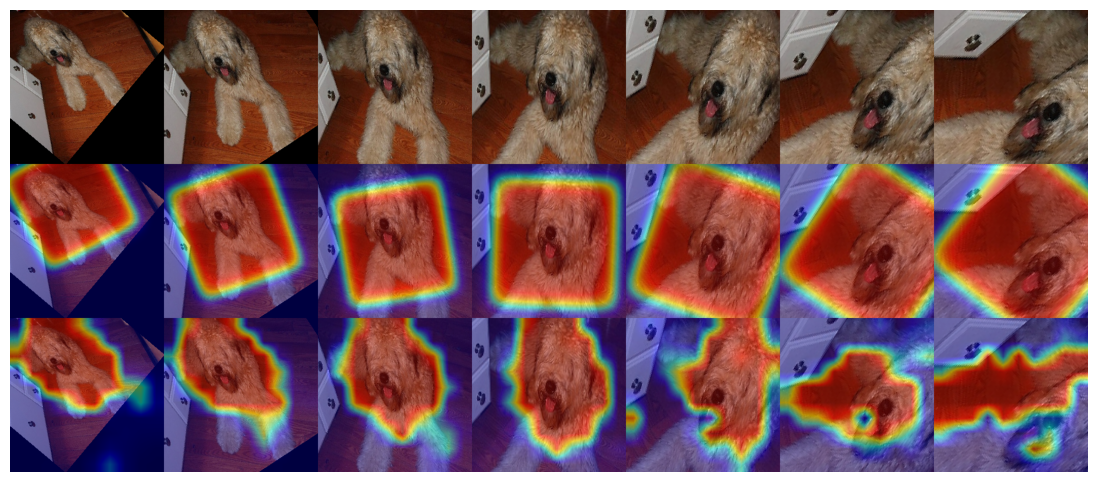}
\end{subfigure}
\caption{Attention maps of our method and position-wise method for a sequence of transforms (composed of rotation, translation and scaling) on each given example of  validation set (Oxford-IIIT Pets). }\label{fig:qualitative_eq}
\end{figure*}

In this section, we qualitatively evaluate the equivariance property of the predicted attention map. For this objective, on a given input, we apply a sequence of transformation, composed of rotation, translation and scaling. Then, we compute the predicted attention map for each of transformed images. The results of our method and position-wise method are shown in Fig. \ref{fig:qualitative_eq}. We see that with our method, the predicted attention maps follow quite smoothly the change along the sequence of transformations. More precisely, it respects quite well the equivariance w.r.t. translation, rotation as well as scaling. In contrast, the position-wise attention maps seem to vary drastically along the transformations. In some cases, for example the last column of Fig. \ref{fig:quali_eq2}, the position-wise method even capture wrongly the interesting part of the image.  This is not surprising as there is not particular reason for this last method to respect the equivariance property. These qualitative results highlight the advantage of our method, in which we can constrain easily 5 rectangle parameters to respect smoothly the equivariance condition.

\subsection{Extended applications: rectangular attention module for descriptive analysis of data}

\begin{figure}[ht]
\centering
\includegraphics[width=0.8\textwidth]{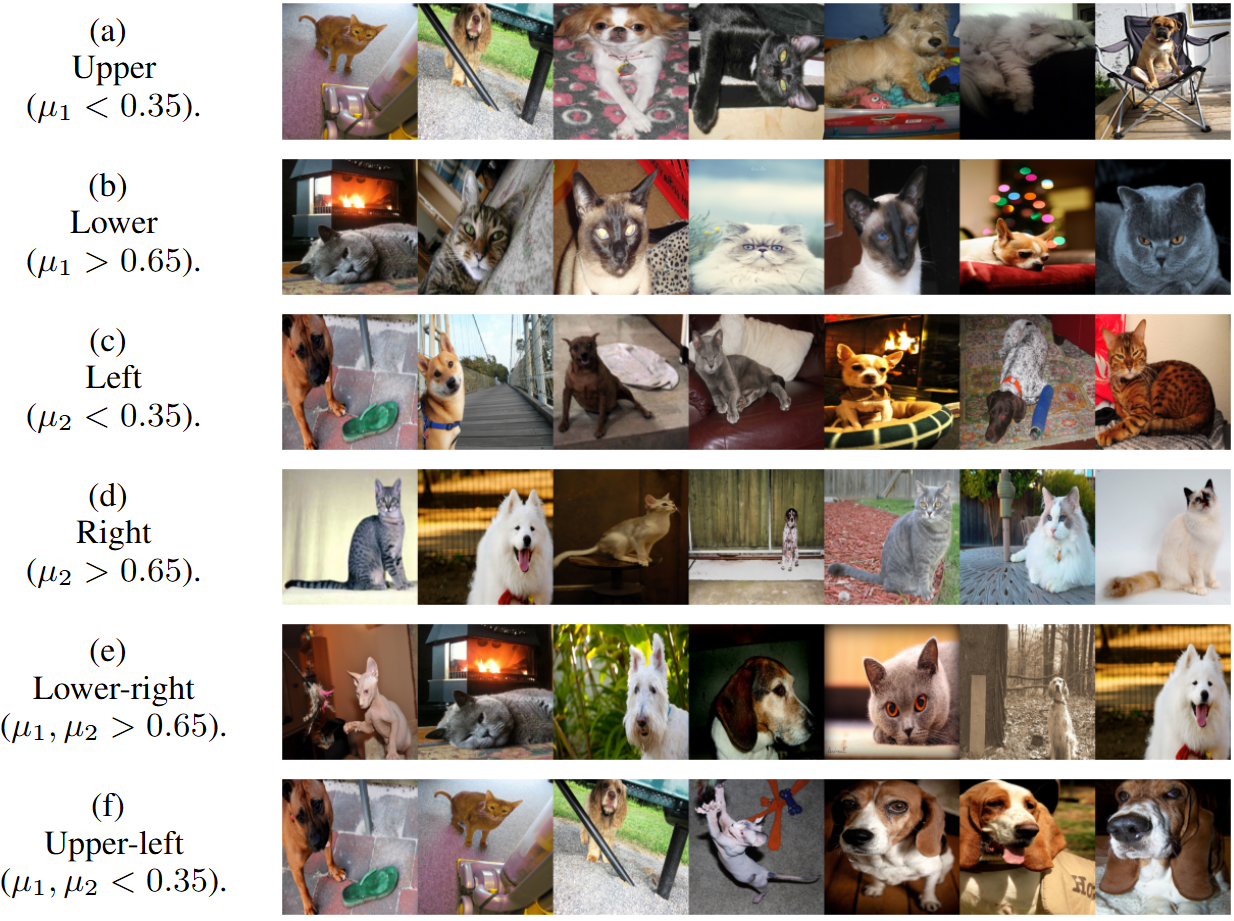}
\caption{Images chosen based on particularity of the center of the predicted rectangle on  validation set (Oxford-IIIT Pets).}\label{fig:descriptive_rtk}
\end{figure}

In some contexts, in addition to having good final predictions, we would also like to better understand the input data. For instance, we may wish to know how objects are positioned in images. In a supervised setting, this can be done using bounding boxes. However, in the the case where we only have at hand object classes but not the labels for the object positions, it is not obvious how we can evaluate this. With our method, we get directly a rectangle that locates the object for each image. As discussed previously in Section \ref{sec:error}, to have a good final prediction for the main model, the predicted rectangle should well capture the object position. Moreover, as each rectangle is characterized by only 5 real scalars, it is convenient to use these parameters to better understand the input data. As a demonstration, we use  the center $(\mu_1,\mu_2)$ of predicted rectangle. If we are looking for images in which the objects are positioned furthest to the left, we can simply choose the images for which $\mu_2$ is small (see Fig. \ref{fig:descriptive_rtk}(c)). Likewise, we can also take the larger values of predicted $\mu_2$ to have the objects positioned furthest to the right (see Fig. \ref{fig:descriptive_rtk}(d)). Proceeding in the same way, we can obtain lower-positioned and upper-positioned objects using predicted $\mu_1$ (see Fig. \ref{fig:descriptive_rtk}(a)-(b)). Furthermore, by using large (resp. small) values for both $\mu_1$ and $\mu_2$, we can also obtain images in which the objects are positioned at the lower-right (resp. the upper-left) corner (Fig. \ref{fig:descriptive_rtk}(e)-(f)). 

Studying the distribution of these statistics gives a better understanding of the data. Note that we get these statistics only by integrating our attention module into an existing base model. Furthermore, we only use the label for classification, and not the label for object positions. This allows us to apply our method to a wider range of applications. Indeed, notating positions (or bounding boxes) requires enormous human resources and time. Finally, note that here we focus only on $(\mu_1,\mu_2)$ to demonstrate how our method can be used as an descriptive analysis tool. However, we can use other parameters such as the predicted orientation $\alpha$ or the predicted $(\sigma_1,\sigma_2)$ to understand the pose as well as the scale of the objects. In conclusion, besides improving model performance and interpretability, our module also acts as a useful tool for a better understanding of the data.

\section{Discussion and Conclusion}

In this paper, we introduce a novel spatial attention mechanism. Its first advantage is its integrability into any convolutional model. It directly provides an attention map in the forward pass. Therefore, it identifies the part of the input image where the main model focuses. This improves interpretability regarding {\it where to look} question.

Moreover, in our experiments, we observe that our method systematically outperforms the position-wise spatial attention module. This is because the shape of our attention map is fixed, helping the model to produce more stable predictions. In contrast, the position-wise method tends to produce irregular boundaries, and thus has more difficulty generalizing to new examples.
Furthermore, we also propose straightforward equivariance constraints to guide the model in generating attention maps. Interestingly, this aligns with self-supervised learning, which enables the model to learn richer features.

For future work, it would be interesting to investigate more diversified shapes beyond the rectangular one while maintaining stability. It could also be worthwhile to study how multiple sub-attention maps might be combined - such as aggregating many small rectangles - to obtain more diverse attention shapes.

\bibliography{main}

\newpage

\appendix

\begin{appendices}

\section{Construct a bell function}\label{sec:construct_bell_function}

The main idea is to construct a differentiable  function with values nearly $1$ inside a rectangle and nearly $0$ outside. 
To build this function, let us first consider the one-dimension case. In this case, we aim to build a function with values nearly $1$ in $[t_0-\sigma,t_0+\sigma]$ and nearly $0$ outside this interval. We will build this function based on the sigmoid function. The sigmoid function is defined as,
\begin{equation}\label{eq:sigmoid_function}
    \Lambda(t) = \frac{e^t}{e^t+1} \ .
\end{equation}

\begin{figure}[ht]

\centering
\includegraphics[width=1.0\textwidth]{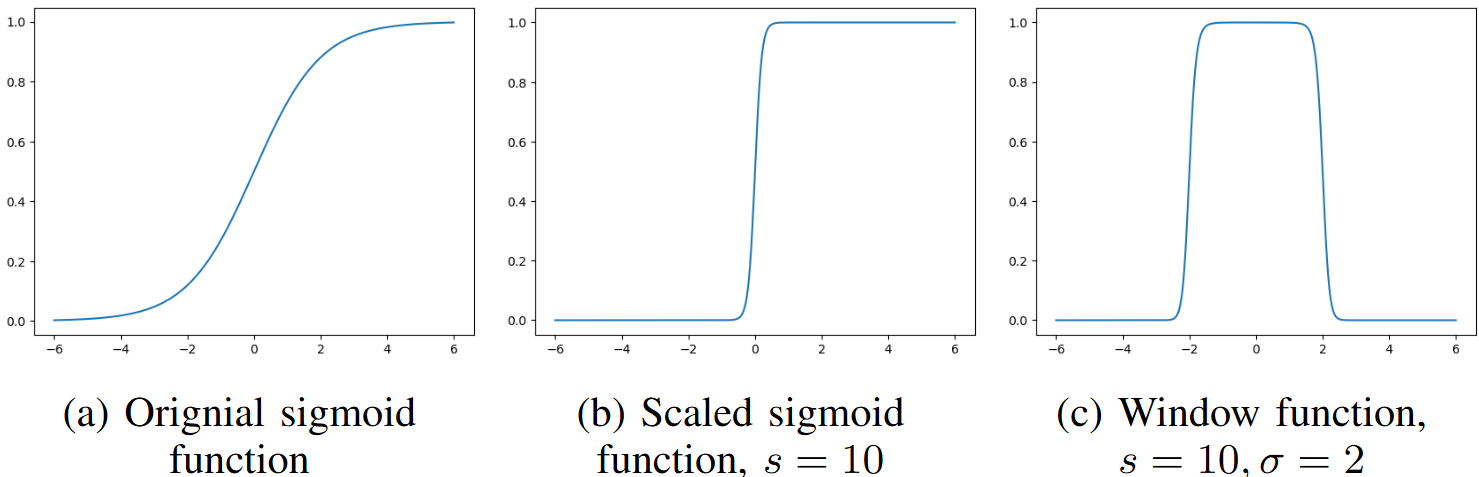}

\caption{Sigmoid function and our window function build from a scaled version of sigmoid function}\label{fig:sigmoid_bell_function_app}
\end{figure}

A well-known property of the sigmoid function is that its value tends quickly to 1 as $t$ tends to $+\infty$ and to $0$  as $t$ tends to $-\infty$ (Fig. \ref{fig:sigmoid_bell_function_app}(a)). Now, to obtain a function that is nearly $1$ with $t>0$ and 0 with $t<0$, we can simply scale by a fixed real number $s>1$:

\begin{equation*}
    \Tilde{\Lambda}_s (t) = \Lambda(s t) = \frac{e^{s t}}{e^{s t}+1} \ .
\end{equation*}

Illustration of scaled sigmoid with $s=10$ is depicted in Fig. \ref{fig:sigmoid_bell_function_app}(b).

Based on this obvious remark, to build a function such that it is nearly $1$ in $[t_0-\sigma,t_0+\sigma]$ and $0$ outside this interval, we can use the following function:

\begin{equation*}
    \hat{\Lambda}_{s,t_0,\sigma}(t) =  \Lambda(s(1-(\frac{t-t_0}{\sigma})^2)) \ .
\end{equation*}

An example of this function is depicted in Fig. \ref{fig:sigmoid_bell_function_app}(c).

\section{Why residual connection in our attention module is a safer choice} \label{sec:res_discuss_app}

Let $\Phi_{\theta}$ denote the part of the main model prior to the attention module, where $\theta$ is the set of parameters of $\Phi$. Recall that $\mathcal{L}_{\text{main}}$ is the loss function for the considered task. Let $x\in\mathbb{R}^{H\times W\times C}$ be an output of $\Phi$. Then $x$ is fed into the attention module to obtain $\hat{x}$ (see Eq. \eqref{eq:attention_function}).

Recall that $f$ generates the attention map $f(x)\in \mathbb{R}^{H\times W}$ and $f(x)\odot x$ is the operation explained in Section \ref{sec:preliminary}. Then $\hat{x}$ is passed through the remaining part of the main model to perform the final prediction.  For $(i,j,c)\in [H]\times[W]\times [C]$, Eq. \eqref{eq:attention_function} rewrites as,
\begin{equation*}
    \hat{x}_{i,j,c} = x_{i,j,c} + f(x)_{i,j} \cdot x_{i,j,c} \ .
\end{equation*}
Therefore, for each parameter $\theta_k \in \theta$, we have,
\begin{equation*}
\frac{\partial \hat{x}_{i,j,c}}{\partial \theta_k} = \left(1+f(x)_{i,j}\right) \cdot \frac{\partial x_{i,j,c}}{\partial \theta_k} + x_{i,j,c}\cdot \frac{\partial f(x)_{i,j}}{\partial \theta_k} \ .
\end{equation*}
Now, according to the chain rule, the partial derivative of the loss function w.r.t $\theta_k$ can be written as,

\begin{equation}\label{eq:attention_res_app}
\begin{split}
\frac{\partial \mathcal{L}_{\text{main}}}{\partial \theta_k} 
&= \sum_{i,j,c} \frac{\partial \mathcal{L}_{\text{main}}}{\partial \hat{x}_{i,j,c}}\cdot \frac{\partial \hat{x}_{i,j,c}}{\partial \theta_k} \\
&= \sum_{i,j,c} \frac{\partial \mathcal{L}_{\text{main}}}{\partial \hat{x}_{i,j,c}} \left(1+f(x)_{i,j}\right) \cdot \frac{\partial x_{i,j,c}}{\partial \theta_k}
+ \sum_{i,j,c} \frac{\partial \mathcal{L}_{\text{main}}}{\partial \hat{x}_{i,j,c}}\cdot x_{i,j,c}\cdot \frac{\partial f(x)_{i,j}}{\partial \theta_k}  \ .    
\end{split}
\end{equation}

If we do not use the residual connection, $\hat{x}_{i,j,c} = f(x)_{i,j} \cdot x_{i,j,c}$, then we have,
\begin{eqnarray*}
\frac{\partial \mathcal{L}_{\text{main}}}{\partial \theta_k} 
= \sum_{i,j,c} \frac{\partial \mathcal{L}_{\text{main}}}{\hat{x}_{i,j,c}} f(x)_{i,j} \cdot \frac{\partial x_{i,j,c}}{\partial \theta_k}
+ \sum_{i,j,c} \frac{\partial \mathcal{L}_{\text{main}}}{\hat{x}_{i,j,c}}\cdot x_{i,j,c}\cdot \frac{\partial f(x)_{i,j}}{\partial \theta_k} \ .
\end{eqnarray*}

Notice that $f(x)$ is only non-zero in a certain support, denoted by $\mathcal{S}$ (see Fig. \ref{fig:ex_rec_attention_intro}). Hence, we have,
\begin{equation}\label{eq:attention_no_res_app} 
\frac{\partial \mathcal{L}_{\text{main}}}{\partial \theta_k}
= \sum_{c} \sum_{(i,j)\in \mathcal{S}} \frac{\partial \mathcal{L}_{\text{main}}}{\hat{x}_{i,j,c}} f(x)_{i,j} \cdot \frac{\partial x_{i,j,c}}{\partial \theta_k}
+ \sum_{i,j,c} \frac{\partial \mathcal{L}_{\text{main}}}{\hat{x}_{i,j,c}}\cdot x_{i,j,c}\cdot \frac{\partial f(x)_{i,j}}{\partial \theta_k}  \ .    
\end{equation}

Let us take a closer look at the second term in this last equation. Note that the attention map $f(x)$ is parametrized by $(\mathbf{\mu}(x), \mathbf{\sigma}(x), \alpha(x))$, which are functions of $x$ themselves. For ease of notation, let us denote $(\mathbf{\mu}, \mathbf{\sigma}, \alpha)$ by $\{\beta_i\}_{i=1...5}$. We have:
$$
\frac{\partial f(x)_{i,j}}{\partial \theta_k}
= \sum_{m=1}^{5} \frac{\partial f(x)_{i,j}}{\partial \beta_m} \cdot \frac{\partial \beta_m}{\partial \theta_k} \ .
$$

The meaning of the second term in Eq. \eqref{eq:attention_no_res_app} is that $(\mathbf{\mu}, \mathbf{\sigma}, \alpha)$ should move in the direction to minimize the loss function (represented by $\frac{\partial f(x)_{i,j}}{\partial \beta_m}$). That is, the support of the attention map is moved in a way to catch up the discriminative part of the input. Then, the moving direction of $(\mathbf{\mu}, \mathbf{\sigma}, \alpha)$ can provide the information to update $\theta$ (represented by $\frac{\partial \beta_m}{\partial \theta_k}$). Hence, the gradient in the second term is very \textit{indirect} in the sense that $\theta$ is updated based on how the support is moved. In the early stage of training, if we have a bad initialization, i.e., the attention map does not cover a good discriminative part of image, then the moving of the support provides very weak information to update $\theta$.

Now, let us consider the first term in Eq. \eqref{eq:attention_no_res_app}. It provides more direct gradients w.r.t. $\theta$ as $x$ is a function of $\theta$. If not using residual connection, the model only uses information inside the support of the attention map to update $\theta$ ($(i,j)\in \mathcal{S}$). Hence, if we have a bad initialization, i.e., the support cover a non-informative part of the image, then we cannot use information outside the support. This makes the optimization more difficult. 

Now, we go back to our setup where we use the residual connection (gradient in Eq. \eqref{eq:attention_res_app}). Comparing Eq. \eqref{eq:attention_res_app} with Eq. \eqref{eq:attention_no_res_app}, we see that Eq. \eqref{eq:attention_res_app} contains a supplementary term, $\sum_{i,j,c} \frac{\partial \mathcal{L}}{\hat{x}_{i,j,c}} \cdot \frac{\partial x_{i,j,c}}{\partial \theta_k}$. This guarantees that we always have the information outside the support of the attention map to update $\theta$. This holds even with a bad initialization, especially in the early stage of the training. This justifies the use of the residual connection in our method.

\section{Proofs of generalization errors} \label{sec:proof_error}
In this section, we provide the complete proofs concerning the generalization errors discussed in Section \ref{sec:error}. 
Our proofs are inspired by the methodology developed in \cite{mohri2018foundations}.
To begin with, we  introduce some notations and results given in \cite{mohri2018foundations}.

\begin{definition}[Empirical Rademacher complexity, p. 30 in \cite{mohri2018foundations}]
Let $H$ be a family of functions mapping from $\mathcal{Z}$ to $[a, b]$ and $S = (z_1,...,z_N)$ a fixed sample of size $N$ with elements in $\mathcal{Z}$. Then, the empirical Rademacher complexity of $H$ with respect to the sample $S$ is defined as:
\begin{equation}
   \widehat{\mathcal{R}}_S(H) = \mathbb{E}_{\sigma}\left[\sup_{h \in H} \frac{1}{N}\sum_{i=1}^{N} \sigma_i h(z_i)\right],
\end{equation}
where $\sigma = (\sigma_1,...,\sigma_N)$, with $\sigma_i$'s independent uniform random variables taking values in $\{-1, +1\}$. The random variables $\sigma_i$ are called Rademacher variables.
\end{definition}

\begin{theorem} [Theorem 3.3, p. 31 in \cite{mohri2018foundations}] \label{theo:general_bound} 
Let $\mathcal{G}$ be a family of functions mapping from $\mathbb{Z}$ to $[0, 1]$. Then, for any
$\delta > 0$, with probability at least $1- \delta$ over the draw of an i.i.d. sample $S$ of size $N$, the following holds for all $g \in \mathcal{G}$:
$$
\mathbb{E}[g(Z)] \leq \frac{1}{N} \sum_{i=1}^{N} g(z_i) + 2\widehat{\mathcal{R}}_S(\mathcal{G}) +3\sqrt{\frac{\log\frac{2}{\delta}}{2N}} \ .
$$

\end{theorem}

Let $\mathcal{G} = \{g: (x,a) \mapsto \Phi(f(x),a), f\in \mathcal{F}\}.$ Recall that $a$ is the ground-truth attention map associated with $x$. We also recall that $\Phi$ is defined in Eq. \eqref{eq:missing_rate}. So, $g$ satisfies that $g(z) \in [0,1], \ \forall z:=(x,a)$. To prove Theorem \ref{theo:missing_rate}, it suffices to prove the following proposition:
\begin{prop}
    The empirical Rademacher complexity of $\mathcal{G}$ w.r.t. $S$ is upper bounded by $\frac{1-\rho}{4}$, i.e., 
    $\widehat{\mathcal{R}}_S(\mathcal{G}) \leq \frac{1-\rho}{4}$.
\end{prop}

\begin{proof}

Let $\sigma = (\sigma_1,...,\sigma_N)$, with $\sigma_i$'s independent uniform random variables taking values in $\{-1, +1\}$. Now we calculate the Rademacher complexity of $\mathcal{G}$. By definition, this term can be can be calculated as,
\begin{equation*}
\begin{split}
\widehat{\mathcal{R}}_S(\mathcal{G}) &= \mathbb{E}_{\sigma}\left[\sup_{g \in \mathcal{G}} \frac{1}{N}\sum_{i=1}^{N} \sigma_i g(z_i)\right] \\
&=  \mathbb{E}_{\sigma}\left[\sup_{f \in \mathcal{F}} \frac{1}{N}\sum_{i=1}^{N} \sigma_i \times \frac{1}{2}\left(1- \Psi(a_i,f(x_i)) \right) \right] \\
&= \mathbb{E}_{\sigma}\left[\sup_{f \in \mathcal{F}} \left( \frac{1}{2N}\sum_{i=1}^{N} \sigma_i - \frac{1}{2N}\sum_{i=1}^{N} \sigma_i\Psi(a_i,f(x_i)) \right) \right] \\
&= \mathbb{E}_{\sigma}\left[\frac{1}{2N}\sum_{i=1}^{N} \sigma_i + \sup_{f \in \mathcal{F}} \left(-\frac{1}{2N}\sum_{i=1}^{N} \sigma_i\Psi(a_i,f(x_i)) \right) \right] \\
&= \mathbb{E}_{\sigma}\left[\frac{1}{2N}\sum_{i=1}^{N} \sigma_i\right] + \mathbb{E}_{\sigma}\left[\sup_{f \in \mathcal{F}} \left(-\frac{1}{2N}\sum_{i=1}^{N} \sigma_i\Psi(a_i,f(x_i)) \right) \right].
\end{split}
\end{equation*}

We note that $\mathbb{E}_{\sigma}\left[\frac{1}{2N}\sum_{i=1}^{N} \sigma_i\right]=\frac{1}{2N}\sum_{i=1}^N \mathbb{E}_{\sigma_i}[\sigma_i]=0$. Thus,
\begin{equation*}
\begin{split}
\widehat{\mathcal{R}}_S(\mathcal{G})
&= \mathbb{E}_{\sigma}\left[\sup_{f \in \mathcal{F}} \left(-\frac{1}{2N}\sum_{i=1}^{N} \sigma_i\Psi(a_i,f(x_i)) \right) \right] \\
&= \mathbb{E}_{\sigma}\left[\sup_{f \in \mathcal{F}} \left(-\frac{1}{2N}\sum_{i=1}^{N} \left(\mathbf{1}_{\{\sigma_i=1\}}-\mathbf{1}_{\{\sigma_i=-1\}}\right)\Psi(a_i,f(x_i)) \right) \right] \\
&= \mathbb{E}_{\sigma}\left[\sup_{f \in \mathcal{F}} \left(\frac{1}{2N}\sum_{i=1}^{N} \left(\mathbf{1}_{\{\sigma_i=-1\}}-\mathbf{1}_{\{\sigma_i=1\}}\right)\Psi(a_i,f(x_i)) \right) \right] \ .
\end{split}
\end{equation*}

Note that $\Psi(a_i,f(x_i)) \leq 1$ and $\mathbf{1}_{\{\sigma_i=-1\}}\geq 0$. Therefore,
$$
\mathbf{1}_{\{\sigma_i=-1\}}\Psi(a_i,f(x_i)) \leq \mathbf{1}_{\{\sigma_i=-1\}} \ . 
$$

So,
\begin{equation*}
\begin{split}
\widehat{\mathcal{R}}_S(\mathcal{G})
&\leq \mathbb{E}_{\sigma}\left[\sup_{f \in \mathcal{F}} \left( \frac{1}{2N}\sum_{i=1}^{N} \left( \mathbf{1}_{\{\sigma_i=-1\}}-\mathbf{1}_{\{\sigma_i=1\}}\times \Psi(a_i,f(x_i)) \right) \right) \right] \\
&= \mathbb{E}_{\sigma}\left[\sup_{f \in \mathcal{F}} \left( \frac{1}{2N}\sum_{i=1}^{N} \mathbf{1}_{\{\sigma_i=-1\}} - \frac{1}{2N}\sum_{i=1}^{N} \left( \mathbf{1}_{\{\sigma_i=1\}}\times \Psi(a_i,f(x_i)) \right) \right) \right] \\
&= \mathbb{E}_{\sigma}\left[ \frac{1}{2N}\sum_{i=1}^{N} \mathbf{1}_{\{\sigma_i=-1\}} + \sup_{f \in \mathcal{F}} \left( - \frac{1}{2N}\sum_{i=1}^{N} \left( \mathbf{1}_{\{\sigma_i=1\}}\times \Psi(a_i,f(x_i)) \right) \right) \right] \\
&= \mathbb{E}_{\sigma}\left[ \frac{1}{2N}\sum_{i=1}^{N} \mathbf{1}_{\{\sigma_i=-1\}} \right]
+ \mathbb{E}_{\sigma}\left[ \sup_{f \in \mathcal{F}} \left( - \frac{1}{2N}\sum_{i=1}^{N} \left( \mathbf{1}_{\{\sigma_i=1\}}\times \Psi(a_i,f(x_i)) \right) \right) \right] \ .
\end{split}
\end{equation*}

Now, note that
\begin{eqnarray*}
\mathbb{E}_{\sigma}\left[ \frac{1}{2N}\sum_{i=1}^{N} \mathbf{1}_{\{\sigma_i=-1\}} \right]
=  \frac{1}{2N}\sum_{i=1}^{N} \mathbb{E}_{\sigma}\left[  \mathbf{1}_{\{\sigma_i=-1\}} \right]
= \frac{1}{2N}\sum_{i=1}^{N} \mathbb{P} (\sigma_i=-1)
= \frac{1}{2N}\sum_{i=1}^{N} \frac{1}{2} = \frac{1}{4}\ .
\end{eqnarray*}
Therefore,
$$
\widehat{\mathcal{R}}_S(\mathcal{G})
\leq \frac{1}{4} + \mathbb{E}_{\sigma}\left[ \sup_{f \in \mathcal{F}} \left( - \frac{1}{2N}\sum_{i=1}^{N} \left( \mathbf{1}_{\{\sigma_i=1\}}\times \Psi(a_i,f(x_i)) \right) \right) \right] \ .
$$

Note that by definition of $\mathcal{F}$, $ \Psi(a_i,f(x_i)) \geq \rho, \ \forall f \in \mathcal{F}$ and $-\mathbf{1}_{\{\sigma_i=1\}}\leq 0$. Hence, 
\begin{eqnarray*}
\widehat{\mathcal{R}}_S(\mathcal{G})
&\leq& \frac{1}{4} + \mathbb{E}_{\sigma}\left[ - \frac{1}{2N}\sum_{i=1}^{N} \rho \mathbf{1}_{\{\sigma_i=1\}} \right] \\
&=&  \frac{1}{4}  - \frac{\rho}{2N}\sum_{i=1}^{N} \mathbb{E}_{\sigma}\left[ \mathbf{1}_{\{\sigma_i=1\}} \right]
= \frac{1}{4}-\frac{\rho}{4} = \frac{1-\rho}{4}\ .    
\end{eqnarray*}

\end{proof}

\section{Proof of Proposition \ref{prop:union_intersection}} \label{sec:proof_union_intersection}

\begin{proof}
First, we consider $\Psi((a(x)+\mathbf{1})/2,(f(x)+\mathbf{1})/2)$.  
\begin{eqnarray*}
\Psi((a(x)+\mathbf{1})/2,(f(x)+\mathbf{1})/2)
&=& \frac{1}{HW}\sum_{i,j}\frac{(a(x)_{i,j}+1)}{2}\cdot \frac{(f(x)_{i,j}+1)}{2}\\
&=& \frac{1}{HW}\sum_{(i,j):a(x)_{i,j}=f(x)_{i,j}=1}\frac{2}{2}\cdot \frac{2}{2} \; (\text{as} \ a(x)_{i,j},f(x)_{i,j}\in \{-1,1\}) \\
&=& \frac{1}{HW}\sum_{(i,j):a(x)_{i,j}=f(x)_{i,j}=1} 1\\
&=& \frac{1}{HW}\left|\{(i,j):a(x)_{i,j}=f(x)_{i,j}=1\}\right|\\
&=& \frac{1}{HW} \left| \mathcal{S}(x)\cap \widehat{\mathcal{S}}(x) \right| \ . 
\end{eqnarray*}

\noindent
Now, we consider $2\Phi((\mathbf{1}-a(x))/2,(\mathbf{1}-f(x))/2)$.

\noindent
As $2\Phi((\mathbf{1}-a(x))/2,(\mathbf{1}-f(x))/2) =  1 - \Psi((\mathbf{1}-a(x))/2,(\mathbf{1}-f(x))/2)$, we now develop $\Psi((\mathbf{1}-a(x))/2,(\mathbf{1}-f(x))/2)$. We have

\begin{eqnarray*}
& & \Psi((\mathbf{1}-a(x))/2,(\mathbf{1}-f(x))/2) \\
&=& \frac{1}{HW}\sum_{i,j}\frac{(1-a(x)_{i,j})}{2}\cdot \frac{1-(f(x)_{i,j})}{2}\\
&=&  \frac{1}{HW}\sum_{(i,j):a(x)_{i,j}=f(x)_{i,j}=-1}\frac{2}{2}\cdot \frac{2}{2} \; (as \  a(x)_{i,j},f(x)_{i,j} \in  \{-1,1\}) \\
&=& \frac{1}{HW}\left|\{(i,j):a(x)_{i,j}=f(x)_{i,j}=-1\}\right| \ . 
\end{eqnarray*}

\noindent
Hence,
\begin{eqnarray*}
2\Phi((\mathbf{1}-a(x))/2,(\mathbf{1}-f(x))/2)
&=& 1 - \Psi((\mathbf{1}-a(x))/2,(\mathbf{1}-f(x))/2)   \\
&=& \frac{1}{HW}\left(HW-\left|\{(i,j):a(x)_{i,j}=f(x)_{i,j}=-1\}\right|\right) \\
&=& \frac{1}{HW} |\{(i,j):a(x)_{i,j}=1 \ \text{or} \ f(x)_{i,j}=1\}| \\
&=& \frac{1}{HW} \left| \mathcal{S}(x)\cup \widehat{\mathcal{S}}(x) \right| \ . 
\end{eqnarray*}

\noindent
So, we have
\begin{equation*}
\frac{\Psi((a(x)+\mathbf{1})/2,(f(x)+\mathbf{1})/2)}{2\Phi((\mathbf{1}-a(x))/2,(\mathbf{1}-f(x))/2)} = \frac{|\mathcal{S}(x)\cap \widehat{\mathcal{S}}(x)|}{|\mathcal{S}(x)\cup \widehat{\mathcal{S}}(x)|} \ .       
\end{equation*}

\end{proof}

\section{Proof of Theorem \ref{theo:divergence_attention}} \label{sec:proof_divergence_attention}
To prove Theorem \ref{theo:divergence_attention}, we first recall the following property (see Chapter 6 in \cite{MR815650}).
\begin{property}\label{cor:tv_distance}
Let $\mathcal{D}_1$ and $\mathcal{D}_2$ (with density $p_1$ and $p_2$ respectively) be probability measures on a measurable space $(\mathcal{X},\sigma)$. Let $p_1$ and $p_2$ be respectively the density of $\mathcal{D}_1$ and $\mathcal{D}_2$ with respect to a reference measure $\nu$. Let $\Gamma_1 = \{x\in \mathcal{X}:p_1(x)\geq p_2(x)\}$ and $\Gamma_2 = \{x\in \mathcal{X}:p_2(x)\geq p_1(x)\}$, we have
\begin{eqnarray*}
\sup_{\Gamma\in\sigma} |\mathcal{D}_1(\Gamma)-\mathcal{D}_2(\Gamma)|
&=& \mathcal{D}_1(\Gamma_1)-\mathcal{D}_2(\Gamma_1)
= \int_{\Gamma_1} (p_1(x)-p_2(x)) \ d\nu(x) \\
&=& \mathcal{D}_2(\Gamma_2)-\mathcal{D}_1(\Gamma_2)
= \int_{\Gamma_2} (p_2(x)-p_1(x)) \ d\nu(x)\ .    
\end{eqnarray*}
That is,
\[
d_{TV}(\mathcal{D}_1, \mathcal{D}_2) = \int_{\Gamma_1} (p_1(x)-p_2(x)) \ d\nu(x)
= \int_{\Gamma_2} (p_2(x)-p_1(x)) \ d\nu(x) \ .   
\]
\end{property}

\noindent
With Property \ref{cor:tv_distance}, we are ready to prove Theorem \ref{theo:divergence_attention}.

\begin{proof}
Consider $\Gamma \in \sigma$. For a pair of classes $(y,y')$, we have
$$
\mathcal{D}_y(\Gamma)-\mathcal{D}_{y'}(\Gamma)
= \int_{\Gamma} \left(p_y(\widehat{x}) - p_{y'}(\widehat{x})\right) \ d\nu(\widehat{x}) \ .
$$
In the following passage, we use $x$ instead of $\mathcal{M}^{-1}(\widehat{x})$ for the sake of brevity.
Note that 
\begin{eqnarray*}
p_y(\widehat{x}) - p_{y'}(\widehat{x})
&=& \widetilde{\Psi}\left(a(x),f(x)\right)\left(p^{\text{pure}}_y(\widehat{x}) - p^{\text{pure}}_{y'}(\widehat{x})\right) \\
&+& (\eta_{y}-\eta_{y'})\left(1-\widetilde{\Psi}\left(a(x),f(x)\right)\right)p_{\text{background}}(\widehat{x}) \ .
\end{eqnarray*}
Hence,
\begin{eqnarray*}
\mathcal{D}_y(\Gamma)-\mathcal{D}_{y'}(\Gamma)
&=&   \int_{\Gamma} \widetilde{\Psi}\left(a(x),f(x)\right)(p^{\text{pure}}_y(\widehat{x})-p^{\text{pure}}_{y'}(\widehat{x})) \  d\nu(\widehat{x})  \\
&+& (\eta_{y}-\eta_{y'})\int_{\Gamma}\left(1-\widetilde{\Psi}\left(a(x),f(x)\right)\right)p_{\text{background}}(\widehat{x}) \ d\nu(\widehat{x}) \ .
\end{eqnarray*}
\noindent
1. Without loss of generality, we assume that $\eta_{y}-\eta_{y'}\geq0$. Let $\Gamma^o = \{x\in\mathcal{X}: p_y^{\text{pure}}(\widehat{x})\geq p_{y'}^{\text{pure}}(\widehat{x})\}$. By the definition of the total variation distance, we have
\begin{eqnarray*}
d_{TV}(\mathcal{D}_y,\mathcal{D}_{y'}) 
&\geq& \mathcal{D}_y(\Gamma^o)-\mathcal{D}_{y'}(\Gamma^o) \\
&=&   \int_{\Gamma^o} \widetilde{\Psi}\left(a(x),f(x)\right)(p^{\text{pure}}_y(\widehat{x})-p^{\text{pure}}_{y'}(\widehat{x})) \  d\nu(\widehat{x}) \\
&+& (\eta_{y}-\eta_{y'})\int_{\Gamma^o}\left(1-\widetilde{\Psi}\left(a(x),f(x)\right)\right)p_{\text{background}}(\widehat{x}) \\
&\geq& \int_{\Gamma^o} \widetilde{\Psi}\left(a(x),f(x)\right)(p^{\text{pure}}_y(\widehat{x})-p^{\text{pure}}_{y'}(\widehat{x})) \  d\nu(\widehat{x})  \\
&\geq& \inf_{x\in\mathcal{X}}\widetilde{\Psi}(a(x),f(x)) \int_{\Gamma^o} (p^{\text{pure}}_y(\widehat{x})-p^{\text{pure}}_{y'}(\widehat{x})) \  d\nu(\widehat{x})  \\
&=& \inf_{x\in\mathcal{X}}\widetilde{\Psi}(a(x),f(x)) \cdot d_{TV}(\mathcal{D}_y^{\text{pure}},\mathcal{D}_{y'}^{\text{pure}}) \ (\text{Property \ref{cor:tv_distance}})\ .
\end{eqnarray*}

In the case where $\eta_{y}-\eta_{y'}\leq 0$, we can have the same result, by using the fact that $d_{TV}(\mathcal{D}_y,\mathcal{D}_{y'}) 
\geq \mathcal{D}_{y'}(\Gamma^o)-\mathcal{D}_{y}(\Gamma^o)$.
So,
$$
d_{TV}(\mathcal{D}_y,\mathcal{D}_{y'}) \geq \inf_{x\in\mathcal{X}}\widetilde{\Psi}(a(x),f(x)) \cdot d_{TV}(\mathcal{D}_y^{\text{pure}},\mathcal{D}_{y'}^{\text{pure}}) \ .
$$

2.  $\mathcal{D}_y^{\text{pure}}$ and $\mathcal{D}_{y'}^{\text{pure}}$ have distinct supports.

Without loss of generality, we assume that $\eta_{y}-\eta_{y'}\geq0$.
Let $\Gamma_y$ be the support of $\mathcal{D}_y$. By the definition of total variation distance, we have
\begin{eqnarray*}
d_{TV}(\mathcal{D}_y,\mathcal{D}_{y'})
&\geq&  \mathcal{D}_y(\Gamma_y)-\mathcal{D}_{y'}(\Gamma_y) \\
&=&  \int_{\Gamma_y} \widetilde{\Psi}\left(a(x),f(x)\right)(p^{\text{pure}}_y(\widehat{x})-p^{\text{pure}}_{y'}(\widehat{x})) \  d\nu(\widehat{x})  \\
&+&  (\eta_{y}-\eta_{y'})\int_{\Gamma^o}\left(1-\widetilde{\Psi}\left(a(x),f(x)\right)\right)p_{\text{background}}(\widehat{x})d\nu(\widehat{x}) \\
&\geq&  \int_{\Gamma_y} \widetilde{\Psi}\left(a(x),f(x)\right)p^{\text{pure}}_y(\widehat{x}) \  d\nu(\widehat{x})   \ (\mathcal{D}_y \ \text{and} \ \mathcal{D}_{y'} \ \text{have distinct supports} ) \\
&=&  \int_{\Gamma_y} \widetilde{\Psi}\left(a(\mathcal{M}^{-1}(\widehat{x})),f(\mathcal{M}^{-1}(\widehat{x})\right)p^{\text{pure}}_y(\widehat{x}) \  d\nu(\widehat{x})  \\
&=&  \mathbb{E}_{\widehat{x}\sim \mathcal{D}_y^{\text{pure}}}\left[\widetilde{\Psi}\left(a(\mathcal{M}^{-1}(\widehat{x})),f(\mathcal{M}^{-1}(\widehat{x})\right)\right] \ .
\end{eqnarray*}
That is,
\begin{eqnarray*}
d_{TV}(\mathcal{D}_y,\mathcal{D}_{y'})
&\geq& \mathbb{E}_{\widehat{x}\sim \mathcal{D}_y^{\text{pure}}}\left[\widetilde{\Psi}\left(a(\mathcal{M}^{-1}(\widehat{x})),f(\mathcal{M}^{-1}(\widehat{x})\right)\right] \\
&\geq& \min_{\widehat{y}\in\{y,y'\}} \left(\mathbb{E}_{\widehat{x}\sim \mathcal{D}_{\widehat{y}}^{\text{pure}}}\left[\widetilde{\Psi}\left(a(\mathcal{M}^{-1}(\widehat{x})),f(\mathcal{M}^{-1}(\widehat{x})\right)\right]\right)
\ .    
\end{eqnarray*}

We can proceed in the same way in the case where $\eta_{y}-\eta_{y'}\leq0$. For conclusion
\begin{eqnarray*}
d_{TV}(\mathcal{D}_y,\mathcal{D}_{y'})
\geq \min_{\widehat{y}\in\{y,y'\}} \left(\mathbb{E}_{\widehat{x}\sim \mathcal{D}_{\widehat{y}}^{\text{pure}}}\left[\widetilde{\Psi}\left(a(\mathcal{M}^{-1}(\widehat{x})),f(\mathcal{M}^{-1}(\widehat{x})\right)\right]\right)
\ .  
\end{eqnarray*}

\end{proof}

\section{Conditions for the attention module to be injective}\label{sec:proof_injective}

First of all, we recall our framework. Attention module $\mathcal{M}$ transforms an $x$ into $\widehat{x}$. That is,

\begin{equation}\label{eq:attention_function_app}
\widehat{x} := \mathcal{M}(x)=x+f(x)\odot x \ . 
\end{equation}
Recall that $x,\widehat{x}\in\mathbb{R}^{H\times W\times C}$ and $f(x)\in\mathbb{R}^{H\times W}$. Eq. \eqref{eq:attention_function_app} means that 
\begin{equation*}
\widehat{x}_{ijk} = x_{ij}+f(x)_{ij}\cdot x_{ijk}, \; \; \forall (i,j,k)\in [H]\times[W]\times[C] \ .
\end{equation*}

\subsection{Conditions for the attention module to be injective with binary simplification}

Firstly, we consider the case with the binary assumption where attention maps are binary-valued.  That is, $f(x),a(x)\in\{0,1\}^{H\times W}$. We place ourselves in the case without rescaling step (in Eq. \eqref{eq:rescale_attention}).

With this simplification, the condition for the attention module to be injective is quite straight-forward and can be stated in the following proposition.
\begin{prop}\label{prop:condition_binary}
    If for all $(x \neq x')\in\mathcal{X}^2\subseteq(\mathbb{R}^{H\times W\times C})^2$, there exists $(i,j,k)\in[H]\times[W]\times[C]$ such that $x_{ijk}\neq x'_{ijk}$ and $f(x)_{ij}=f(x')_{ij}$, then $\mathcal{M}$ is injective.
\end{prop}

\begin{proof}
By definition, $\widehat{x}_{ijk}=x_{ijk}+f(x)_{ij}\cdot x_{ijk}$ and $\widehat{x'}_{ijk}=x'_{ijk}+f(x')_{ij}\cdot x'_{ijk}$. Using assumption $x_{ijk}\neq x'_{ijk}$ and $f(x)_{ij}=f(x')_{ij}$, we have that $\widehat{x}_{ijk}\neq \widehat{x'}_{ijk}$. Hence $\widehat{x}\neq \widehat{x'}$ by definition.

That is, $x \neq x'$ implies $\mathcal{M}(x)\neq \mathcal{M}(x')$. As this holds true for any $x \neq x'$, $\mathcal{M}$ is injective. This completes the proof.
\end{proof}

\begin{remark}
\begin{itemize}
    \item The condition in Proposition \ref{prop:condition_binary} is reasonable, as for two different inputs $x$ and $x'$, it suffices to have a position $(i,j)$ and a channel $k$ such that the input elements are different ($x_{ijk}\neq x'_{ijk}$) and receive the same predicted attention value at that position $(i,j)$ (either $0$ or $1$).
    \item In contrast, a necessary condition for $\mathcal{M}$ to be not injective is that, for two different inputs $x$ and $x'$, at all positions $(i,j)$ and channels $k$ where $x_{ijk}\neq x'_{ijk}$, they have different predicted attention values ($f(x)_{ij}\neq f(x')_{ij}$). 
\end{itemize}
\end{remark}

\subsection{Conditions for the attention module to be injective without binary simplification}

Now, we consider the general case where we do not make the binary assumption. That is, $f(x),a(x)\in[0,1]^{H\times W}$. In this case, the condition of injectivity is less straight-forward that shall be presented in this section. 

We assume that there exists a $\delta>0$ such that $\|x\|\leq\delta,\; \forall x\in\mathcal{X}$. That is, $\mathcal{X}$ is included in the hyper-sphere $S(0,\delta)$. Then, in the case without rescaling step (in Eq. \eqref{eq:rescale_attention}), the condition for $\mathcal{M}$ to be injective can be stated as follows.

\begin{prop}
If for all $(x \neq x')\in S(0,\delta)^2\subseteq(\mathbb{R}^{H\times W\times C})^2$, \[\mathlarger{\sum}_{i}^{H}\mathlarger{\sum}_{j}^{W}\mathlarger{\sum}_{k}^{C}\left(f(x)_{ij}\cdot x_{ijk} -f(x')_{ij}\cdot x'_{ijk} \right)^2<\mathlarger{\sum}_{i}^{H}\mathlarger{\sum}_{j}^{W}\mathlarger{\sum}_{k}^{C}\left(x_{ijk} - x'_{ijk} \right)^2 \ , \]
then $\mathcal{M}$ is injective. 
\end{prop}

In the case where we apply rescaling step (in Eq. \eqref{eq:rescale_attention}), the result above holds true by simply replacing $f$ by $\widetilde{f}$.

To prove this proposition, we introduce the following theorem needed for the proof.


\begin{theorem}[Chapter 5, p.122 in \cite{ortega1970iterative}]\label{theorem:contraction}
 Suppose that $A$ is a non-singular linear operator from $\mathbb{R}^n$ to $\mathbb{R}^n$  and that \( G: D \subset \mathbb{R}^n \to \mathbb{R}^n \)  is a mapping such that in a closed ball \( S_0 = S(x^0, \delta) \subset D \),
\[
\|Gx - Gy\| \leq \alpha \|x - y\|, \quad \forall x, y \in S_0,
\]
where $0 < \alpha < \beta^{-1}, \quad \beta = \|A^{-1}\|.$ Then the mapping \( F: S_0 \to \mathbb{R}^n \), defined by \( F x = A x - G x \), \( x \in S_0 \),
is a homeomorphism between \( S_0 \) and \( F(S_0) \).
\end{theorem}

Now, we are ready for the proof.

\begin{proof}
We have that $\mathcal{M}(x)=\widehat{x}=x+f(x)\odot x$. Hence we can write $\mathcal{M}$ as 
$$
\mathcal{M}(x) = \mathbf{I}x+ H(x) = \mathbf{I}x - \left(-H(x)\right) \ ,
$$
where $H(x) = f(x)\odot x$. Now, we apply Theorem \ref{theorem:contraction} with $A=\mathbf{I}$ and $G=-H$. To prove the proposition, it suffices to show that $-H$ is contractive. This is equivalent to prove $H$ is contractive. Notice that $\|H(x)-H(x')\|^2_2 = \mathlarger{\sum}_{i}^{H}\mathlarger{\sum}_{j}^{W}\mathlarger{\sum}_{k}^{C}\left(f(x)_{ij}\cdot x_{ijk} -f(x')_{ij}\cdot x'_{ijk} \right)^2$ and $\|x-x'\|^2_2 = \mathlarger{\sum}_{i}^{H}\mathlarger{\sum}_{j}^{W}\mathlarger{\sum}_{k}^{C}\left(x_{ijk} - x'_{ijk} \right)^2$. Hence, using the condition in the proposition, we have that $H$ is contractive. This completes the proof.
\end{proof}

\section{Experiment Details}\label{sec:exp_details}

For all the experiments, we train both models EfficientNet-b0 \cite{tan2019efficientnet} and MobileNetV3 \cite{howard2019searching} for 40 epochs (we used models pretrained (on ImageNet) as initialization). We use {\it Adam} optimizer with a learning rate of $10^{-4}$. For both models, we add a linear layer with hidden dimension of $512$ before the softmax layer (with output dimension of $37$, which is the number of classes).

During training, we apply data augmentation techniques including {\it ColorJitter} (brightness=0.3, contrast=0.3, saturation=0.3) and {\it RandomPerspective}. We use batch size equal to 32 for training.

\textbf{Our attention module.} The attention module is composed of 3 convolutional layers with output channel dimensions of 64, 128 and 256 respectively. Each layer has a kernel size equal to 3, stride equal to 1, padding equal to 1. After each of the first two convolutional layers in the attention module, we apply the Max Pooling with kernel size of 2, which allows to reduce the spatial dimension 4 times after each layer. Then, we apply the Global Averange Pooling (GAP) along the spatial dimension after the third layer, to obtain the feature vector of 256. Finally, this is fed to a linear layer that has  output dimension of 5, which is the number of rectangle parameters. 

\textbf{Position-wise attention module}. We follow the design proposed in the original work of \cite{park2018bam}. The attention module is composed of 4 convolutional layers. The first three layers have output channel dimension of 80. The last layer has the output channel dimension of 1 (as the output of this layer is the attention map). The first layer has kernel size of 1, playing the role of dimension reduction. The other layers have kernel size equal to 3, dilation equal to 4 (dilated convolution), padding equal to 4. All layers have stride equal to 1. Notice that with this method, one cannot apply sub-sampling techniques over spatial dimension, to maintain the spatial dimension. Hence we do not increase the channel dimension after each layer, as this can lead to significant increase in running time. Besides, the dilated convolution is used to have a large receptive field, as discussed in \cite{park2018bam}.

\textbf{Complexity.} We notice that with our method, after each layer, the spatial dimension is reduced by 4 times after each layer, but we only increase the channel dimension by 2 times after each layer. Regarding the position-wise module, the spatial dimension and channel dimension is maintained after each layer. Hence, our method is approximately twice smaller than position-wise module, if not to mention that our feature dimension is 64, which is smaller than 80 in position-wise module.

For EfficientNet-b0, we insert the attention module after the layer \textbf{model.\_blocks[15]}. For MobileNetV3, we insert the attention module after passing by \textbf{model.features($\cdot$)}.

\end{appendices}

\end{document}